\RequirePackage{fix-cm}
\documentclass[smallextended]{svjour3}       % onecolumn (second format)
\smartqed  % flush right qed marks, e.g. at end of proof
\usepackage{graphicx}
\usepackage{lineno,hyperref}
\usepackage{amsmath}
\usepackage{multirow}
\usepackage[misc]{ifsym}
\usepackage{amssymb}
\usepackage{latexsym}
\usepackage{url}
\usepackage{xcolor}
\usepackage{algorithm}
\usepackage{algpseudocode}
\usepackage{subfigure}

\begin{document}

\title{Grassmannian Graph-attentional Landmark Selection for Domain Adaptation%\thanks{Grants or other notes
}


\author{Bin Sun \and Shaofan Wang  \and  Dehui Kong    \and Jinghua~Li  \and  Baocai~Yin
           %etc.
}

%\authorrunning{Bin Sun         \and Dehui Kong  \and Shaofan Wang   \and Lichun Wang \and Yuping Wang   \and Baocai Yin} % if too long for running head

\institute{Bin Sun \and Shaofan Wang  \and  Dehui Kong    \and Jinghua~Li  \and  Baocai~Yin \at
              Beijing Key Laboratory of Multimedia and Intelligent Software Technology, BJUT Faculty of Information Technology, Beijing University of Technology, Beijing 100124, China \\
              \email{binsun008@gmail.com}             \\
           Shaofan Wang (\Letter) \at
             Tel.: +86 188 1307 8200 \\
             \email{wangshaofan@bjut.edu.cn}
}

\date{Received: date / Accepted: date}

\maketitle

\begin{abstract}
Domain adaptation aims to leverage information from the source domain to improve the classification performance in the target domain. It mainly utilizes two schemes: sample reweighting and feature matching.
While the first scheme allocates different weights to individual samples, the second scheme matches the feature of two domains using global structural statistics. The two schemes are complementary with each other, which are expected to jointly work for robust domain adaptation. {Several methods combine the two schemes, but the underlying relationship of samples is insufficiently analyzed} due to the neglect of the hierarchy of samples and the geometric properties between samples. To better combine the advantages of the two schemes, we propose a Grassmannian graph-attentional landmark selection (GGLS) framework for domain adaptation. {GGLS presents a landmark selection scheme using attention-induced neighbors of the graphical structure of samples and performs distribution adaptation and knowledge adaptation over Grassmann manifold.
the former treats the landmarks of each sample differently, and the latter avoids feature distortion and achieves better geometric properties.} Experimental results on different real-world cross-domain visual recognition tasks demonstrate that GGLS provides better classification accuracies compared with state-of-the-art domain adaptation methods.
\end{abstract}

\section{Introduction}
Images, videos, and other multimedia data are growing rapidly in the big data era.
Annotating precise labels to massive unlabeled data, which is pervasive and important for supervised learning, remains a tedious and inaccurate task, since it is time-consuming and depends on different subjective judgments of experts. To solve this problem, what we should notice is that the distributions of training data (source domain) and testing data (target
domain) are different. Therefore, How to reduce the distribution difference of different domains is the key to solve this problem.
Domain adaptation~\cite{pan2010survey,cui2014flowing,qian2017cross,ding2015deep,ma2019deep} provides an effective means to settle this issue, which is referred to as leveraging label information from the source domain to improve the classification performance in the target domain by transfer learning.
Domain adaptation has wide applications in many research topics such as video retrieval~\cite{yang2016metric,zhang2016cross}, image classification~\cite{daras2011search,zheng2019multiple}, objection recognition~\cite{gopalan2011domain,niu2016domain}, and text categorization~\cite{long2015domain}.

The aim of domain adaptation is to explore common latent information and thus reduce the difference of distribution between the source domain and the target domain. Existing methods can be roughly grouped into two categories, namely, sample reweighting and feature matching.

Sample reweighting methods~\cite{aljundi2015landmarks}~\cite{hubert2016learning}~\cite{xu2017unified} mainly reweight the samples from the source domain or both domain and select the most relevant samples, i.e., landmarks, to match the distribution of the target domain. However, most of the methods either simply eliminate some abnormal samples or simply select landmarks that are common to all samples, which are not select independent landmarks what is appropriate for each sample.

Feature matching methods~\cite{pan2010domain}~\cite{gong2012geodesic}~\cite{ghifary2017scatter}~\cite{wang2018visual} extract a robust feature for the samples from the source domain and the target domain, which consider either the distribution adaptation to reduce the difference of marginal and conditional distributions between domains or important data properties, e.g., statistical property and geometric structure. However, most of the methods only consider the distribution differences of two domains, which do not consider the processing of the samples that are not conducive to feature matching, which can easily lead to negative transfer.

In summary, the above methods consider sample reweighting and feature extraction independently, which respectively reduce the distribution divergence on two different scales by selecting the most relevant samples and matching the feature of two domains using global structural statistics. In fact, these two types of methods are complementary with each other and
the performance will be better by exploring them simultaneously.
Some methods~\cite{long2014transfer}~\cite{li2018transfer} have indicated that the combination of the two schemes works and is expected to serve as an effective solution for solving the issues of the aforementioned methods. However, the methods~\cite{long2014transfer}~\cite{li2018transfer} simply focus on the combination of the two sample reweighting and feature extraction, and still only consider the landmarks are the same for all the samples. Besides, these methods mainly extract features in the original space, which are prone to feature distortion. In other words, these methods neglect the hierarchy of samples and the geometric properties between samples.

To address these problems, we also combine feature matching with instance reweighting, and propose Grassmannian graph-attentional landmark selection (GGLS) framework for robust domain adaptation. Technically, GGLS is formulated from three aspects: 1) GGLS analyzes global characteristics of samples and perform distribution adaptation, which optimizes a nonparametric maximum mean discrepancy (MMD) using the weighted contribution of marginal and conditional distributions, which introduce a balanced parameter to evaluate the relative contribution of the two distributions; 2) GGLS analyzes local and individual characteristics of samples and perform knowledge adaptation, which transfers important information including the local topology structure information of samples and the discrimination information of samples from source domain; 3) GGLS analyzes abnormal samples and perform feature selection. Furthermore, to avoid feature distortion and have better geometric properties, our method is to learn over the Grassmann manifold. For another, GGLS presents a landmark selection scheme using a graph-attentional mechanism, which maintains the hierarchy of samples by treating the landmarks of each sample differently in domain adaptation.
Experimental results show that GGLS achieves better performance when compared with other state-of-the-art methods including traditional domain adaptation methods and depth domain adaptation methods.

The main contributions of our work are summarized as follows.
\begin{enumerate}
	\item We propose a domain adaptation framework, which integrates sample reweighting and feature matching.
	\item We propose a graph-attentional mechanism based sample reweighting, which treats the landmarks of each sample differently.
	\item We jointly learn distribution adaptation and knowledge adaptation in Grassmann manifold, which can avoid feature distortion and have better geometric properties.
\end{enumerate}

The remainder of this paper is organized as follows. Section~\ref{Related Work} introduces the related work. Section~\ref{Methodology} first elaborates on the primary model based on distribution adaptation and knowledge adaptation, and then elaborates on some improvements based on primary model including Grassmannian manifold feature learning, feature selection, and graph-attentional landmark selection. Section~\ref{Analysis} introduces some analysis. We evaluate the performance of the proposed method by the experimental results in Section~\ref{Experiments}. The conclusion is given in Section~\ref{Conclusion}.
\section{Related Work}
\label{Related Work}
In this section, we briefly review the work related to our proposed method. We review the domain adaptation methods, which can be roughly grouped into two categories: sample
reweighting and feature matching.

\subsection{Sample reweighting methods}
The sample reweighting method aims to reduce the distribution difference by reweighting the samples according to the correlation between the source domain and the target domain.
Dai et al.~\cite{dai2007boosting} adjusted the weights of the samples from the source domain to filter out the samples, which are most unlikely drown from the distribution of the target domain. The reweighted samples from the source domain compose a distribution similar to the one found at the target domain.
Wan et al.~\cite{wan2011bi} proposed a Bi-weighting domain adaptation method, which adjusts the samples and feature weights of the training data from the source domain.
Aljundi et al.~\cite{aljundi2015landmarks} proposed a landmarks selection extracted from both domains so as to reduce the distribution difference between the source and target domains, and then non linearly project the data in the same space where an efficient subspace alignment is performed.
Hubert et al.~\cite{hubert2016learning} learned representative landmarks with the ability to identify the adaptation ability of each sample with a properly assigned weight, and the adaptation capabilities of such cross-domain landmarks can be determined accordingly, which can achieve promising performance.
Xu et al.~\cite{xu2017unified} learned the instance weights, which bridges the distributions of different domains, and Mahalanobis distance is learned to constrain intra-class and inter-class distances for the target domain, which make knowledge transfer across domains more effective.
Li et al.~\cite{li2018transfer} proposed a method for landmark selection based on graph, which is transformed to select the vertices with a high degree. The method focuses on the samples, and neatly sidesteps the sorts of computational issues raised by massive matrix operations.

\subsection{Feature matching methods}
The feature matching method aims to conduct subspace learning by using the matching of subspace geometric structure or feature distribution, so as to reduce the difference of marginal distribution or conditional distribution between the source domain and the target domain.
This kind of method mainly process features. A better feature can make the model obtain excellent performance~\cite{wu2015service}. From the subsequent experimental results in Table~\ref{Office+Caltech10_SURF},~\ref{Office+Caltech10_DeCaf}, we can see the influence of different features on the model. In general, features need to be further processed, so that features have a better discriminant ability. Some feature processing techniques include feature selection~\cite{li2019dividing}, sparse representation~\cite{xu2015discriminative}, feature sampling~\cite{chen2019matting},  feature preservation~\cite{liang20203d}, structural consistency~\cite{hou2016unsupervised}, distribution matching~\cite{cao2018unsupervised} and so on. These techniques can also be applied to the field of domain adaptation.
Pan et al.~\cite{pan2010domain} proposed the transfer component analysis to learn transfer components across domains considering marginal distribution shift.
In the subspace spanned by these transfer components, data distributions in different domains are close to each other.
On the basis of the work of ~\cite{pan2010domain}, Long et al.~\cite{long2013transfer} added the conditional distribution in a principled dimensionality reduction procedure, and constructed a new feature that is effective for substantial distribution difference.
Many methods were improved based on~\cite{long2013transfer}. Sun et al.~\cite{sun2016return} aligned the second-order statistics of distributions of source and target domains, without requiring any target labels.
Xu et al.~\cite{xu2015discriminative} used a sparse matrix to model the noise, which is more robust to different types of noise.
Tahmoresnezhad et al.~\cite{tahmoresnezhad2017visual} constructed condensed domain invariant clusters in the embedding representation to separate various classes alongside the domain transfer.
Fernando et al.~\cite{fernando2013unsupervised} learned a linear mapping in a new subspace that aligns the source domain with the target domain.
Sun et al.~\cite{sun2015subspace} extended the work of~\cite{fernando2013unsupervised} by adding the subspace variance adaptation.
Ghifary et al.~\cite{ghifary2017scatter} proposed scatter component analysis based on the simple geometrical measure, which minimizing the mismatch between domains, and maximizing the separability of data, each of which is quantified through scatter.
Zhang et al.~\cite{zhang2017joint} learned two projections to project the source and target data into respective subspaces, which reduces the shift between domains both statistically and geometrically.
Gong et al.~\cite{gong2012geodesic} proposed to learn the geodesic flow kernel between domains in the manifold. The method models domain shift by integrating an infinite number of subspaces that characterize changes in geometric and statistical properties from the source to the target domain.
Wang et al.~\cite{wang2018visual} learned a classifier in Grassmann manifold with structural risk minimization while performed dynamic distribution alignment, which can solve the problems of both degenerated feature transformation and unevaluated distribution alignment.
{Chen et al.~\cite{chen2019joint} proposed to boost the transfer performance by jointing domain alignment and discriminative feature learning.
Ganin et al~\cite{ganin2016domain} introduced domain-adversarial neural network by adding domain-adversarial loss, which helps to learn more transferable and discriminative features. Following the idea, many work adopted domain-adversarial training~\cite{zhang2018collaborative},~\cite{kumar2018co}.
Roy et al.~\cite{roy2019unsupervised} proposed domain alignment layers that implement feature whitening for matching source and target feature distributions.
Wang et al.~\cite{wang2019unifying} proposed a confidence-aware pseudo label selection scheme to gradually align the domains in an iterative learning strategy.
Wang et al.~\cite{wang2020unsupervised} proposed a selective pseudo-labeling approach by incorporating supervised subspace learning and structured prediction based pseudo-labeling into an iterative learning framework. 
Chen et al.~\cite{chen2019progressive} proposed a Network by exploiting the intra-class variation in the target domain to align the discriminative features across domains progressively and effectively.
}

Unlike most of the above methods, following the work~\cite{long2014transfer}~\cite{li2018transfer}, we also simultaneously perform feature matching and instance reweighting, and propose GGLS. GGLS comprehensively analyzes global, local and individual characteristics of
sample data for feature matching, and also eliminates abnormal samples which are not conducive to domain adaptation. Furthermore, GGLS performs Grassmannian manifold feature learning for better geometric property description and graph-attentional mechanism, which treat the landmarks of each sample differently in domain adaptation.

\section{Methodology}
\label{Methodology}
\subsection{Notations}
In this paper, we use nonbold letters, bold lowercase letters and bold uppercase letters to denote scalars, vectors and matrices, respectively. $\mathbf{I}$ represents the identity matrix, $||\cdot||_{2}$ denotes the $\ell_{2}$ norm of a matrix, $||\cdot||_{F}$ denotes the Frobenious norm of a matrix, $||\cdot||_{2,1}$ denotes the $\ell_{2,1}$ norm of a matrix (i.e. the sum of $\ell_2$ norm of all rows of a matrix) and $tr(\cdot)$ denotes the trace operation.
\subsection{Problem Definition}

Denote the labeled data of source domain to be
\begin{align}
\mathcal{D}_{s}=\{\textbf{\textit{x}}_{s}^{i}, l_{s}^{i}\}_{i=1}^{N_{s}}=\{\textbf{\textit{X}}_{s},\textbf{\textit{l}}_{s}\}
\nonumber
\end{align}
and unlabeled data of target domain to be
\begin{align}
\mathcal{D}_{t}=\{\textbf{\textit{x}}_{t}^{i}\}_{i=1}^{N_{t}}=\{\textbf{\textit{X}}_{t}\}
\nonumber
\end{align}
where $\textbf{\textit{X}}_{s}\in\mathbb{R}^{d\times N_{s}}$,
$\textbf{\textit{X}}_{t} \in\mathbb{R}^{d\times N_{t}} $, $\textbf{\textit{l}}_{s}=[l_{s}^{1},\ldots,l_{s}^{N_{s}}]^{\top}\in\mathbb{R}^{N_{s}}$ is data label of source domain, $d$ is the dimension of feature, $N_{s}$ and $N_{t}$ are the sample number of the source domain and the target domain, respectively.
Assume the feature spaces and label spaces between domains are the same: $\mathcal{X}_{s}=\mathcal{X}_{t}$ and $\mathcal{Y}_{s}=\mathcal{Y}_{t}$, while the marginal distribution and the conditional distribution are not same: $P(\textbf{\textit{X}}_{s})\neq P(\textbf{\textit{X}}_{t})$ and $P(\textbf{\textit{l}}_{s}|\textbf{\textit{X}}_{s})\neq P(\textbf{\textit{l}}_{t}|\textbf{\textit{X}}_{t})$, where $\textbf{\textit{l}}_{t}=[l_{t}^{1},\ldots,l_{t}^{N_{t}}]^{\top}\in\mathbb{R}^{N_{t}}$ is data label of target domain.

\subsection{Primary Model}
The key to domain adaptation is to directly or indirectly mine the connections between the source domain and the target domain, since these connections are the bridges to implement information transfer. To explore these connections, we need to consider the following first: 1) If there is no alignment between the source domain and the target domain, the source domain and the target domain will not be able to connect, thus the information in the source domain will be difficult to transfer to the target domain; 2) For a classification problem, the most important information to transfer is the discriminant information; 3) Abnormal samples need to be processed.
As a result, it is natural to learn the transformation matrix $\mathbf{W}\in\mathbb{R}^{D\times C}$ using the following primary model:
\begin{align}\label{formalization}
\min_{\mathbf{W}}\mathcal{L}_{DA}(\mathbf{W},\textbf{\textit{X}})+
\mathcal{L}_{KA}(\mathbf{W},\textbf{\textit{X}},\textbf{\textit{H}}_{s})+\Omega(\mathbf{W})
\end{align}
where $\mathcal{L}_{DA}$ and $\mathcal{L}_{KA}$ represent distribution adaptation (\textrm{DA}) term and knowledge adaptation (\textrm{KA}) term, respectively, $\Omega$ is regularization term, $\textbf{\textit{X}}=[\textbf{\textit{X}}_{s},\textbf{\textit{X}}_{t}]\in\mathbb{R}^{D\times (N_{s}+N_{t})}$ is the feature matrix, $\textbf{\textit{H}}_s=[\textbf{\textit{h}}_{s}^{1},\ldots,\textbf{\textit{h}}_{s}^{N_{s}}]\in\mathbb{R}^{C \times N_{s}}$ is the ground truth label matrix of the source domain, and $\textbf{\textit{h}}_{s}^{n}=[0,\ldots,1,\ldots,0]^{\top}\in\mathbb{R}^C$ is the label vector of the $n$th data of the source domain whose $c$th element is one and others vanish. In the rest of the section, we will present the details of $\mathcal{L}_{DA}$, $\mathcal{L}_{KA}$ and $\Omega$.

\subsubsection{Distribution Adaptation}
The key to distribution adaptation is to minimize the distribution difference between domains. Most existing methods~\cite{long2013transfer,zhang2017joint} minimize the difference in marginal distribution and conditional distribution between domains with equal weight, which is not realistic. Therefore, we introduce a balanced parameter $\mu$ to evaluate the relative contribution of the two distributions. We employ the empirical Maximum Mean Discrepancy (MMD)~\cite{ben2007analysis} to calculate the difference of distributions. The objective function is as follows:
\begin{align}
&\mathcal{L}_{DA}=(1 - \mu )
\|\frac{\sum_{\textbf{\textit{x}}^{i}\in \textbf{\textit{X}}_s}\mathbf{W}^{\top}\textbf{\textit{x}}^{i}}{N_s}-\frac{\sum_{\textbf{\textit{x}}^{i}\in {\textbf{\textit{X}}_t}}\mathbf{W}^{\top}\textbf{\textit{x}}^{i}}{N_t}\|_2^2 \nonumber \\
&~~~~~~~+\mu \sum_{c=1}^{C}||\frac{\sum_{\textbf{\textit{x}}^{i}\in {\textbf{\textit{X}}_{s,c}}}\mathbf{W}^{\top}\textbf{\textit{x}}^{i}}{N_{s,c}}-\frac{\sum_{\textbf{\textit{x}}^{i}\in {\textbf{\textit{X}}_{t,c}}}\mathbf{W}^{\top}\textbf{\textit{x}}^{i}}{N_{t,c}}\|_2^2 \nonumber\\
&~~~~=tr(\mathbf{W}^{\top}\textbf{\textit{X}}\mathbf{M}\textbf{\textit{X}}^{\top}\mathbf{W}) \label{Distribution Adaptation}
\end{align}
where $\mathbf{M}=(1 - \mu ) \mathbf{M}_{0} + \mu \sum_{c = 1}^C {\mathbf{M}_{c}}$, $\mathbf{M}_{0}$ and $\mathbf{M}_{c}$ are given by:
\begin{align}\label{Mr0}
{[{\mathbf{M}_{0}}]_{i,j}} =& \left\{ \begin{array}{ll}
\frac{1}{N_{s}^2},  & \mbox{if~}  {\textbf{\textit{x}}^{i}},{\textbf{\textit{x}}^{j}} \in \textbf{\textit{X}}_{s}\\
\frac{1}{{{N_{t}^2}}},& \mbox{if~}{\textbf{\textit{x}}^{i}},{\textbf{\textit{x}}^{j}} \in \textbf{\textit{X}}_{t}\\
-\frac{1}{{ N_{s}N_{t}}},&  \text{otherwise}
\end{array} \right.  \\
{[{\mathbf{M}_{c}}]_{i,j}} =& \left\{\begin{array}{{l}}
\frac{1}{N_{s,c}^2},~\mbox{if~} {\textbf{\textit{x}}^{i}},{\textbf{\textit{x}}^{j}} \in \textbf{\textit{X}}_{s,c}\\
\frac{1}{N_{t,c}^2},~\mbox{if~} {\textbf{\textit{x}}^{i}},{\textbf{\textit{x}}^{j}} \in \textbf{\textit{X}}_{t,c} \\
-\frac{1}{{N_{s,c}N_{t,c}}},\left\{ \begin{array}{{l}}
\!\!{\mbox{if~}\textbf{\textit{x}}^{i}\in \textbf{\textit{X}}_{s,c},\textbf{\textit{x}}^{j}\in \textbf{\textit{X}}_{t,c}}\\
\!\!{\mbox{if~}\textbf{\textit{x}}^{i}\in \textbf{\textit{X}}_{t,c},\textbf{\textit{x}}^{j}\in \textbf{\textit{X}}_{s,c}}
\end{array} \right.\\
0,~\text{otherwise}
\end{array} \right.
\label{Mrc}
\end{align}
respectively, where $N_{s,c}$ ($N_{t,c}$, resp.) is the sample number of the $c$th class from the source (target, resp.) domain,
$\textbf{\textit{X}}_{s,c}$ ($\textbf{\textit{X}}_{t,c}$, resp.) is the feature matrix of the $c$th class from the source (target, resp.) domain. Since the data of the target domain is unlabeled, we use a classifier trained by the samples of the source domain to predict the pseudo labels of samples of the target domain. In consideration of the less reliability of the pseudo labels, we iteratively update the prediction value.
$\mu$ can be calculated by:
\begin{align}\label{mu}
\mu  = 1 - \frac{tr(\mathbf{W}^{\top}\textbf{\textit{X}}\mathbf{M_{0}}\textbf{\textit{X}}^{\top}\mathbf{W})}{tr(\mathbf{W}^{\top}\textbf{\textit{X}}\mathbf{M}\textbf{\textit{X}}^{\top}\mathbf{W})}
\end{align}
%The smaller $\mu$ is, the larger the difference of samples from the source and the target domains is, the more important the marginal distribution adaptation is; and vice versa.

\subsubsection{Knowledge Adaptation}
Distribution adaptation just focuses on all the data of the source domain and the target domain, and fails to discover the local geometrical structure of all the data and discrimination information of each sample from the source domain, which is more important in discriminative analysis.

In general, a sample tends to share the same label with its k-nearest neighbors~\cite{yan2006graph,li2018transfer}. Thus, it would be helpful if we can make samples of the same class very close during the distribution adaptation. In order to preserve the local geometrical structure of samples from two domains, we construct the graph Laplacian. Let $\mathbf{G}\in \mathbb{R}^{N\times N}$ be similarity matrix with $[\mathbf{G}]_{ij}$ characterizes the favorite relationship among samples. We minimize the following target function:
\begin{align}\label{locality_preserving_projection}
&\frac{1}{2}\sum_{i,j = 1}^{N} {{{\| {{\textbf{W}^{\top}}{\textbf{\textit{x}}_{i}} - {\textbf{W}^{\top}}{\textbf{\textit{x}}_{j}}} \|}_{2}^2}} [\textbf{G}]_{ij} \nonumber \\
=&\sum_{i=1}^{N} \mathbf{W}^{\top} \textbf{\textit{x}}_{i} [\mathbf{S}]_{ii} \textbf{\textit{x}}_{i}^{\top} \mathbf{W}-\sum_{i, j=1}^{N} \mathbf{W}^{\top} \textbf{\textit{x}}_{i} [\mathbf{G}]_{ij} \textbf{\textit{x}}_{j}^{\top} \mathbf{W} \nonumber \\
=&tr(\mathbf{W}^{\top}\textbf{\textit{X}}\mathbf{L}\textbf{\textit{X}}^{\top}\mathbf{W})
\end{align}
where $N=N_{s}+N_{t}$, $\mathbf{L}=\mathbf{S}-\mathbf{G}$, and $\mathbf{S}$ is diagonal matrix with $[\mathbf{S}]_{ii}=\sum_{j=1}^{N}[\mathbf{G}]_{ij}$.

Furthermore, we use label information from the source domain to constrain the new features from the source domain to be discriminative. Therefore, the knowledge adaptation term is as follows:
\begin{equation}
\label{Knowledge Adaptation}
\mathcal{L}_{KA}\!=\!\beta tr(\mathbf{W}^{\top}\textbf{\textit{X}}\mathbf{L}\textbf{\textit{X}}^{\top}\mathbf{W})
\!+\! \gamma\| {{\textbf{\textit{H}}_s} \!-\! \mathbf{W}^{\top}\textbf{\textit{X}}_{s}\|_F^2}
\end{equation}
where the first term penalizes the original locality using graph Laplacian, and the second term penalizes the discrimination power using labels of source domain.
$\beta$ and $\gamma$ are the regularization parameters.
\subsubsection{Overall Formulation}
To prevent overfitting, we add a constraint term $\|\mathbf{W}\|_{F}$. Therefore, we combine Eqs.(\ref{Distribution Adaptation}), (\ref{Knowledge Adaptation}) and regularization term together, and formulate the primary model as follws:
\begin{align}\label{primary_model}
\min_{\mathbf{W}}~&tr(\mathbf{W}^{\top}\textbf{\textit{X}}\mathbf{M}\textbf{\textit{X}}^{\top}\mathbf{W})+\!\beta tr(\mathbf{W}^{\top}\textbf{\textit{X}}\mathbf{L}\textbf{\textit{X}}^{\top}\mathbf{W})\! \nonumber\\
&+\! \gamma\|\textbf{\textit{H}}_s-\mathbf{W}^{\top}\textbf{\textit{X}}_{s}\|_F^2+\lambda_1\|\mathbf{W}\|_{F}^{2}
\end{align}
where $\lambda_1$ is the regularization parameter.
\subsection{Improvement Strategy of Model~(\ref{primary_model})}

\subsubsection{Grassmannian Manifold Feature Learning}
In general, the manifold can be regarded as a low dimensional ``smooth" surface embedded in a high dimensional Euclidean space. The manifold can be approximated as Euclidean space in a small local neighborhood. When extracting the features of samples with nonlinear structure, the manifold representation method achieves good results. Since high dimensional multimedia data exhibit potential geometric properties as they locate in an intrinsic manifold, which can avoid feature distortion. Therefore, we want to transform the features in the original space into manifolds. Among many manifolds, feature transformation and distribution matching in Grassmann manifold usually have effective numerical forms, so they can be expressed and solved efficiently in the transfer learning problem. Therefore, transfer learning on Grassmann manifold is feasible. In this paper, we use Geodesic Flow Kernel (GFK) in~\cite{gong2012geodesic} to transform the features in the original space to Grassmann manifold. We state only the main idea of GFK, and the details can be found in~\cite{gong2012geodesic}.

Denote ${{\mathbf{S}}_s},{{\mathbf{S}}_t}\in{\mathbb{R}^{D \times d}}$ to be the subspaces of the source and target domains, $\mathbf{R}_{S}\in \mathbb{R}^{D \times(D-d)}$ to be the orthogonal complement to ${\mathbf{S}}_s$, namely $\mathbf{R}_{s}^{\top} \mathbf{S}_{s}=\mathbf{0}$, where $D$ is the dimension of the data and $d$ is the subspace dimension. $\mathbb{G}(d,D)$ can be regarded as a collection of all $d$-dimensional subspaces, and the original subspace of each $D$-dimension can be regarded as a point in $\mathbb{G}(d,D)$.
Using the canonical Euclidean metric for Grassmann manifold, the geodesic flow is parameterized as:
\begin{align}
{\bf{\Phi }}:t \in [0,1] \to {\bf{\Phi }}(t) \in \mathbb{G}(d,D)
\nonumber
\end{align}
under the constraints $ {\bf{\Phi }}(0)= {\mathbf{S}}_s $ and  $ {\bf{\Phi }}(1)= {\mathbf{S}_t} $. $\mathbf{\Phi}(t)$ can be represented as follows:
\begin{align}\label{Manifold1}
\mathbf{\Phi}(t)=\mathbf{S}_s \mathbf{U}_{1} \mathbf{\Gamma}(t)-\mathbf{R}_{s} \mathbf{U}_{2} \mathbf{\Sigma}(t)
\end{align}
where $\mathbf{U}_{1} \in \mathbb{R}^{d \times d}$ and $ \mathbf{U}_{2} \in \mathbb{R}^{(D- d) \times d} $are orthogonal matrices. They are given by the following pair of SVDs:
\begin{align}\label{Manifold2}
&\mathbf{S}_{s}^{\top} \mathbf{S}_{t}=\mathbf{U}_{1} \mathbf{\Gamma} \mathbf{V}^{\top}, \nonumber \\
&\mathbf{R}_{s}^{\top} \mathbf{S}_{t}=-\mathbf{U}_{2} \mathbf{\Sigma} \mathbf{V}^{\top}
\end{align}
where both $\mathbf{\Gamma}$ and $\mathbf{\Sigma}$ are $d\times d$ diagonal matrices.
We construct the geodesic flow between the two points and integrate an infinite number of subspaces along the flow. Specifically, the original features are projected into a subspace of infinite dimensions, thereby reducing the phenomenon of domain drift. GFK can be seen as an incremental "walking" approach from ${\bf{\Phi }}(0)$ to ${\bf{\Phi }}(1)$. Then, the features in Grassmann manifold can be expressed as ${\bf{z}} = {\bf{\Phi }}{(t)^{\top}}{\bf{x}}$, and the inner product of two features is expressed as:
\begin{equation}\label{GFK}
\langle{\mathbf{z}_i},{\mathbf{z}_j}\rangle =\int_0^1 {(\mathbf{\Phi }{{(t)}^{\top}}{{\textbf{\textit{x}}}_i)}^{\top}} ({\mathbf{\Phi }}{(t)^{\top}}{{\textbf{\textit{x}}}_j})dt = {{\textbf{\textit{x}}}_i}^{\top}{\mathbf{Q}}{{\textbf{\textit{x}}}_j}
\end{equation}
where $\mathbf{Q}$ is a semi-positive definite matrix. Therefore, the new feature in Grassmann manifold can be represented as  ${\mathbf{z}} = {\mathbf{\Phi }}{(t)^{\top}}{\textbf{\textit{x}}}=\mathbf{Q}^{\frac12}\textbf{\textit{x}}$.

\subsubsection{Kernelization}
To match both the first-order and high-order statistics, we propose to perform the work in the reproducing kernel Hilbert space (RKHS). Considering the kernel mapping
\begin{align}
\psi : \mathbf{z} \rightarrow \psi(\mathbf{z}), \psi(\mathbf{Z})=\left[\psi\left(\mathbf{z}_{1}\right), \ldots, \psi\left(\mathbf{z}_{N}\right)\right]
\nonumber
\end{align}
and the kernel matrix
\begin{align}\label{kernel}
\mathbf{K}=[\mathbf{k}_{1},\ldots,\mathbf{k}_{N}]=\psi(\mathbf{Z})^{\top} \psi(\mathbf{Z})\in \mathbb{R}^{N\times N}
\end{align}
We use the Representer theorem $\mathbf{W}=\psi(\mathbf{Z}) \mathbf{P}$ to kernelize PCA, where $\mathbf{P}\in \mathbb{R}^{N\times C}$.

\subsubsection{Feature Selection}
Distribution adaptation and knowledge adaptation comprehensively analyze global, local and individual characteristics of samples. However, there are always exist some samples that have interference effects. Therefore, we propose a feature selection scheme to handle these abnormal samples, which can more conducive to the adaptation. We introduce $\ell_{2,1}$ norm for transformation matrix $\mathbf{P}$, which leads to a row-sparse regularization. Since each row of $ \mathbf{P}\in \mathbb{R}^{N\times C}$ corresponds to a sample, the row-sparsity can effectively eliminate some abnormal samples.

Based on these, we reformulate our model~(\ref{primary_model}) as follows
\begin{align}\label{second_model}
\min_{\mathbf{P}} ~&tr(\mathbf{P}^{\top}\mathbf{K}(\mathbf{M}+\beta\mathbf{L})\mathbf{K}\mathbf{P})+\gamma\|\textbf{\textit{H}}_s-\mathbf{P}^{\top}\mathbf{K}\|_F^2
\nonumber \\
&+\lambda_1\|\mathbf{P}\|_{2,1}+\lambda_2tr(\mathbf{P}^{\top}\mathbf{K}\mathbf{P})
\end{align}
where $\lambda_1$ and $\lambda_2$ are the regularization parameters. From Eq.(\ref{locality_preserving_projection}), we know that $\mathbf{L}=\mathbf{S}-\mathbf{G}$. In the RKHS, similarity matrix $[\mathbf{G}]_{ij}$ can be represented as
\begin{equation}\label{G_RKHS}
[\textbf{G}]_{ij}=
\begin{cases}
\frac{\textbf{z}_{i}^{\top}\textbf{z}_{j}}{||\textbf{z}_{i}||_{2}||\textbf{z}_{j}||_{2}},&\text{if $\textbf{z}_{i}\in\mathcal{N}_k(\textbf{z}_{j})$~or~$\textbf{z}_{j}\in\mathcal{N}_k(\textbf{z}_{i})$ }\\
0,&\text{otherwise}
\end{cases}
\end{equation}
where $\mathcal{N}_k(\mathbf{z})$ represents the set of the $k$ nearest neighbors of $\mathbf{z}$ in the manifold. In this case, we can use knowledge adaptation term to preserve the
local geometrical structure of nearest points in the manifold.

\subsection{Graph-attentional Landmark Selection}
\begin{figure}
	\centering
	\includegraphics[scale=1]{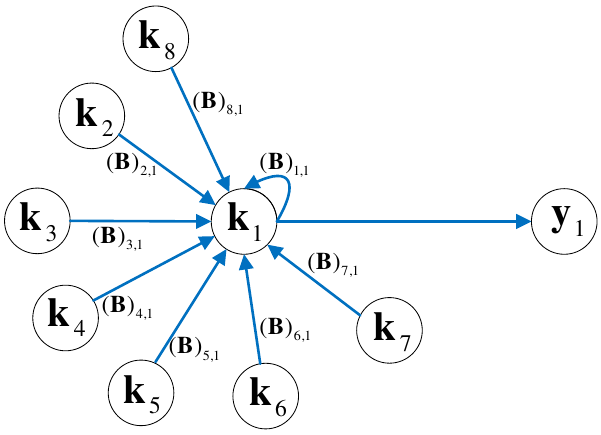}
	\caption{ The aggregation process of graph attention.}
	\label{attention_graph}
\end{figure}

The feature selection scheme can effectively eliminate some abnormal samples, but the relationship between the samples is not analyzed. Sample reweighting based methods~\cite{aljundi2015landmarks}~\cite{long2014transfer}~\cite{hubert2016learning} can solve the problem effectively. These methods performed knowledge transfer using landmarks, i.e., the most relevant samples, which is optimal than using all the samples. However, these methods assume that the landmarks of samples are the same, i.e., the set of the most relevant samples is uniform with respect to each sample. However, the assumption ignores the hierarchy of samples. Motivated by the advantage of attention mechanism~\cite{velickovic2017graph}, we introduce a graph-attentional mechanism based landmark selection, whose idea is to represent each point in the graph by its neighbors using attention coefficient matrix. Through assigning arbitrary weights to neighbors, it can be applied to graph nodes with varying degrees.
In order to obtain features with sufficient expression, a shared transformation matrix $\mathbf{P}$ is applied to each sample .
We use cosine similarity to compute the attentional coefficient matrix $\mathbf{A}$ as follows:
\begin{equation}\label{B_attention coefficient}
[\mathbf{A}]_{ij}=
\begin{cases}
\frac{(\mathbf{P}^{\top}\mathbf{k}_{i})^{\top} \mathbf{P}^{\top}\mathbf{k}_{j}}{\|\mathbf{P}^{\top}\mathbf{k}_{i}\|_2 \|\mathbf{P}^{\top}\mathbf{k}_{j}\|_2},&\text{if $\mathbf{P}^{\top}\mathbf{k}_{i}\in\mathcal{N}_k(\mathbf{P}^{\top}\mathbf{k}_{j})$ }\\
0,&\text{otherwise}
\end{cases}
\end{equation}
which indicates the importance of feature of sample $i$ to sample $j$, where
$\mathcal{N}_k(\textbf{\textit{X}})$ represents the set of $k$ nearest neighbors of $\textbf{\textit{X}}$.
During the optimization, we normalize them across all choices of $j$:
\begin{equation}\label{normalize}
[\mathbf{B}]_{ij}=\frac{[\mathbf{A}]_{ij}}{\sum_{k\in\mathcal{N}_j}[\mathbf{A}]_{kj}}
\end{equation}
We then use $\mathbf{B}$ to calculate a linear combination of the features corresponding to them, and the final output features of each sample can be represented by:
\begin{equation}
\mathbf{y}_{j}=\sum_{i\in\mathcal{N}_j} \mathbf{P}^{\top}\mathbf{k}_{i}(\mathbf{B})_{i,j}= \mathbf{P}^{\top}\mathbf{K}\mathbf{b}_{j}
\end{equation}
where $\mathbf{b}_{j}$ represents the $j$th column of $\mathbf{B}$.
The aggregation process of the graph-attentional mechanism is illustrated in Figure~\ref{attention_graph}.
Feature selection and landmark selection analyze the features of the samples from different aspects. The former eliminates some abnormal samples that are not conducive to adaptation, which is analyzed from the perspective of global samples, and the latter represents each sample by its neighbors, which is analyzed from the perspective of local samples. The combination of the two schemes can effectively promote domain adaptation.

We add graph-attentional landmark selection to the model (\ref{second_model}),and the final model is as follows:
\begin{align}\label{final_model}
\min_{\mathbf{P}} &~tr(\mathbf{P}^{\top}\mathbf{K}\mathbf{B}(\mathbf{M}+\beta\mathbf{L})\mathbf{B}^{\top}\mathbf{K}\mathbf{P})+\gamma\|(\textbf{\textit{H}}-\mathbf{P}^{\top}\mathbf{K}\mathbf{B})\mathbf{R}\|_F^2 \nonumber \\
&+\lambda_1\|\mathbf{P}\|_{2,1}+\lambda_2tr(\mathbf{P}^{\top}\mathbf{K}\mathbf{P})
\end{align}
where $\lambda_1$ and $\lambda _2$ are the regularization parameters,  $\mathbf{R}\in\mathbb{R}^{N\times N}$ is a diagonal domain indicator matrix with $\mathbf{R}_{ii} = 1$ if $i$ belongs to source domain, otherwise $\mathbf{R}_{ii} = 0$, $\textbf{\textit{H}}$ is the ground truth label matrix of samples from source and target domains, where the labels of samples from target domain can be filtered out by the diagonal matrix $\mathbf{R}$.

\subsection{Optimization}
For the optimization problem (\ref{final_model}), we take the derivative with respect to $\mathbf{P}$, set the derivative to zero, and obtain:
\begin{equation}\label{solve_P}
\mathbf{P} =\gamma(\mathbf{KB}\mathbf{U}\mathbf{B}^\top\mathbf{K}+\lambda_{1}\mathbf{F}+\lambda_2\mathbf{K})^{-1}
\mathbf{KBR}\textbf{\textit{H}}^{\top}
\end{equation}
where $\mathbf{U}=\mathbf{M}+\beta\mathbf{L}+\gamma\mathbf{R}$. Note that $\|\mathbf{P}\|_{2,1}$ is not smooth, so we compute its subgradient $\mathbf{F}$~\cite{gu2011joint}~\cite{li2016joint} as follows:
\begin{equation}\label{solve_F}
\mathbf{F}_{i,i}=
\begin{cases}
0,&\text{if $\mathbf{p}^{i}=\mathbf{0}$}\\
\frac{1}{2\|\mathbf{p}^{i}\|_2},&\text{otherwise}
\end{cases}
\end{equation}
where $\mathbf{p}^{i}$ denotes the $i$th row of $\mathbf{P}$. Then, we compute $\mathbf{B}$ via Eq.~(\ref{B_attention coefficient})(\ref{normalize}).
We show an algorithm of solving~(\ref{final_model}) in Algorithm~\ref{algorithm1}, {and give algorithm for testing in Algorithm~\ref{algorithm2}.}
\begin{algorithm}
	\caption{ Optimization of Eq.~(\ref{final_model})}
	\label{algorithm1}
	\begin{algorithmic}[1]
		\Require
		Data $\textbf{\textit{X}}=[\textbf{\textit{X}}_{s}, \textbf{\textit{X}}_{t}]$; label matrix $\textbf{\textit{H}}_{s}$; manifold feature dimension $d$; parameters $\beta, \gamma, \lambda_{1}, \lambda_{2}$;
		\Ensure Transformation matrix $\mathbf{P}$, attention coefficient matrix $\mathbf{B}$;
		\State Initialize $\mathbf{A}=\mathbf{I}$, $\mathbf{F}=\mathbf{I}$;
		\State Compute $\mathbf{B}$ via Eq.~(\ref{normalize});
		\State Learn manifold feature transformation kernel $\mathbf{Q}$ via Eq.~(\ref{GFK}), and compute manifold feature ${\mathbf{Z}}=\mathbf{Q}^\frac12\textbf{\textit{X}}$;
		\State Compute kernel matrix $\mathbf{K}=\Psi(\mathbf{Z},\mathbf{Z})$ via Eq.~(\ref{kernel});
		\State Obtain the pseudo labels of $\mathbf{K}_{t}$ via classifier trained by $\mathbf{K}_{s}$;
		\While {not converge or reach maximum iterations}
		\State Compute $\mathbf{M}$ via Eq.~(\ref{Mr0})(\ref{Mrc});
		\State Compute $\mu$ via Eq.~(\ref{mu});
		\State Compute $\mathbf{P}$ via Eq.~(\ref{solve_P});
		\State Compute $\mathbf{F}$ via Eq.~(\ref{solve_F});
		\State Compute $\mathbf{B}$ via Eq.~(\ref{B_attention coefficient})(\ref{normalize});
		\State Compute the pseudo labels of $\mathbf{Y}_{t}$ via classifier trained by $\mathbf{Y}_{s}$;
		\EndWhile
	\end{algorithmic}
\end{algorithm}

\begin{algorithm}
	\caption{ Testing of GGLS}
	\label{algorithm2}
	\begin{algorithmic}[1]
		\Require
		{Data $ \textbf{\textit{X}}_{t}$; manifold feature $\mathbf{Z}_{s}$;manifold feature transformation kernel $\mathbf{Q}$; Transformation matrix $\mathbf{P}$, attention coefficient matrix $\mathbf{B}$;}
		{\Ensure Labels of $\mathbf{Y}_{t}$;}
		\State {Compute manifold feature ${\mathbf{Z}_{t}}=\mathbf{Q}^\frac12\textbf{\textit{X}}_{t}$;}
		\State {Compute kernel matrix $\mathbf{K}=\Psi([\mathbf{Z}_{s},\mathbf{Z}_{t}],[\mathbf{Z}_{s},\mathbf{Z}_{t}])$ via Eq.~(\ref{kernel});}
		\State {Compute output feature $\mathbf{Y}_{t}= \mathbf{P}^{\top}\mathbf{K}\mathbf{B}_{N_{s}+1:N}$;}
		\State {Compute the labels of $\mathbf{Y}_{t}$ via classifier trained by $\mathbf{Y}_{s}$;}
	\end{algorithmic}
\end{algorithm}

	{\section{Analysis}
		\label{Analysis}
	\subsection{Theory Analysis}}	
	{In general, our final model is presented in the coarse-to-fine form. We first propose a coarse model (\ref{primary_model}) by analysis as a whole, and then optimize the model for some specific problems to further improve the model performance.  
	To minimize the difference of distribution between the source domain and target domain, our coarse model (\ref{primary_model}) analyzes global, local, and individual characteristics of sample data. According to the three characteristics, we respectively propose distribution adaptation term (the first term of model (\ref{primary_model}), a locality preservation term (the second term of model (\ref{primary_model}), and a discrimination preservation term (the third term of model \ref{primary_model}). The three terms can balance the contributions of marginal and conditional distributions based on all the samples, preserve the local topology structure of samples from source and target domains, and preserve the discriminative power of samples from the source domain, respectively. To better exploit the three characteristics of model \ref{primary_model}), we analyze it from three aspects. First, We transform the sample features in the original space into manifolds, which can avoid feature distortion. Secondly, we propose feature selection scheme to handle these abnormal samples. Thirdly, we analyze the relationship between the samples by landmark selection. In fact, these three aspects are mainly discussed from the three perspectives of feature space, feature selection and the relationship between features. It can be seen from the subsequent experiments that our model has significantly improved after further improvement.}

	{\subsection{Complexity Analysis}
	The computational costs of Algorithm~\ref{algorithm1} consists of five major parts:
	computing $\mathbf{B}$ via solving Eq.~(\ref{normalize}),
	computing $\mathbf{K}$ via Eq.~(\ref{kernel}),
	computing $\mathbf{M}$ via Eq.~(\ref{Mr0})(\ref{Mrc}),
	computing $\mathbf{P}$ via Eq.~(\ref{solve_P}),
	and computing $\mathbf{F}$ via Eq.~(\ref{solve_F}). The computational complexity can be seen in Table~\ref{tab-time}.
	The computation of $\mathbf{B}$ costs $\mathcal{O}$($N^{2}$),
	the computation of   $\mathbf{K}$ costs $\mathcal{O}$($N^{2}$),
	the computation of $\mathbf{M}$  costs $\mathcal{O}$($N^{2}$),
	the computation of $\mathbf{P}$ costs $\mathcal{O}$($N^{2}$),
	the computation of $\mathbf{F}$ costs $\mathcal{O}$($N^{2}$).
	Suppose the Algorithm~\ref{algorithm1} will converge after $T$ iterations, the overall computational cost of Algorithm~\ref{algorithm1} is $\mathcal{O}(5TN^{2})$.
}
\begin{table}[t] 
	\caption{{Phase-wise computational complexity.}\label{tab-time}}
	\centering
	%\resizebox{!}{9.9mm}
	{
		\begin{tabular}{ccc}
			\hline
			Phase  & Line num of Algorithm 1 & Complexity \\
			\hline
			Computation of $\mathbf{B}$ &line~2  &$\mathcal{O}$($N^{2}$) \\
			Computation of $\mathbf{K}$ &line~4  &$\mathcal{O}$($N^{2}$)	\\
			Computation of $\mathbf{M}$ &line~8&$\mathcal{O}$($N^{2}$) \\
			Computation of $\mathbf{P}$ &line~9&$\mathcal{O}$($N^{2}$) \\
			Computation of $\mathbf{F}$	&line~10  &$\mathcal{O}$($N^{2}$)	\\
			\hline
		\end{tabular}
	}
\end{table}

\subsection{Convergence Analysis}
\label{Convergence Analysis}

Algorithm~\ref{algorithm1} solves problem~(\ref{final_model}) with locally optimal solution of $\mathbf{P}$. In order to demonstrate this convergence, $\emph{Lemma 3}$ is utilized
subsequently, which is proposed and proven in~\cite{nie2010efficient}.

$\emph{Lemma 3}$: For any nonzero vectors $\mathbf{u}, \mathbf{v} \in \mathbb{R}^{c}$, it holds that
\begin{align}\label{Lemma_model}
\|\mathbf{u}\|_{2}-\frac{\|\mathbf{u}\|_{2}^{2}}{2\|\mathbf{v}\|_{2}} \leq\|\mathbf{v}\|_{2}-\frac{\|\mathbf{v}\|_{2}^{2}}{2\|\mathbf{v}\|_{2}}
\end{align}

$\emph{Theorem 1}$: Algorithm~\ref{algorithm1} decreases problem~(\ref{final_model}) by iteratively updating $\mathbf{P}$ with its optimal solution until convergence.

$\emph{Proof}$: Denote the objective value in the $t$th iteration of problem~(\ref{final_model}) as $\emph{f}(\mathbf{P}_{(t)})$, i.e.,
\begin{align}\label{Proof1}
\emph{f}(\mathbf{P}_{(t)})=tr(\mathbf{P}_{(t)}^{\top}\mathbf{\Gamma}\mathbf{P}_{(t)})+\gamma\|(\textbf{\textit{H}}-\mathbf{P}_{(t)}^{\top}\mathbf{K}\mathbf{B})\mathbf{R}\|_F^2
\end{align}
where $ \mathbf{\Gamma}=\mathbf{K}\mathbf{B}(\mathbf{M}+\beta\mathbf{L})\mathbf{B}^{\top}\mathbf{K}+\lambda_1\mathbf{F}+\lambda_2\mathbf{K}$.

Since Algorithm~\ref{algorithm1} updates $\mathbf{P}$ with the optimal solution in each iteration, for the $(t + 1)$th iteration, it must hold
\begin{align}\label{Proof2}
\emph{f}(\mathbf{P}_{(t+1)}) \leq \emph{f}(\mathbf{P}_{(t)})
\end{align}
and
\begin{align}\label{Proof22}
\emph{f}(\mathbf{P}_{(t+1)})+\lambda_{1} \sum_{i=1}^{N} \frac{\|\mathbf{p}_{(t+1)}^{i}\|^{2}}{2\|\mathbf{p}_{(t)}^{i}\|} \leq
\emph{f}(\mathbf{P}_{(t)})+\lambda_{1} \sum_{i=1}^{N} \frac{\|\mathbf{p}_{(t)}^{i}\|^{2}}{2\|\mathbf{p}_{(t)}^{i}\|}
\end{align}
According to $\emph{Lemma 3}$, we have
\begin{align}\label{Proof3}
\begin{array}{l}{\|\mathbf{p}_{(t+1)}^{i}\|_{2}-\frac{\|\mathbf{p}_{(t+1)}^{i}\|_{2}^{2}}{2\|\mathbf{p}_{(t)}^{i}\|_{2}}}
{\quad \leq \|\mathbf{p}_{(t)}^{i}\|_{2}-\frac{\|\mathbf{p}_{(t)}^{i}\|_{2}^{2}}{2\|\mathbf{p}_{(t)}^{i}\|_{2}}} \\
{\Rightarrow \lambda_{1} \sum_{i=1}^{N}\|\mathbf{p}_{(t+1)}^{i}\|_{2}-\lambda_{1} \sum_{i=1}^{N} \frac{\|\mathbf{p}_{(t+1)}^{i}\|_{2}^{2}}{2\|\mathbf{p}_{(t)}^{i}\|_{2}}} \\
{\quad \leq \lambda_{1} \sum_{i=1}^{N}\|\mathbf{p}_{(t)}^{i}\|_{2}-\lambda_{1} \sum_{i=1}^{N} \frac{\|\mathbf{p}_{(t)}^{i}\|_{2}^{2}}{2\|\mathbf{p}_{(t)}^{i}\|_{2}}}\end{array}
\end{align}
Then, adding~(\ref{Proof3}) to~(\ref{Proof22}), we have
\begin{small}
	\begin{align}\label{Proof4}
	&\emph{f}(\mathbf{P}_{(t+1)})+\lambda_{1} \sum_{i=1}^{N} \frac{\|\mathbf{p}_{(t+1)}^{i}\|^{2}}{2\|\mathbf{p}_{(t)}^{i}\|}
	+\lambda_{1}\sum_{i=1}^{N}(\|\mathbf{p}_{(t+1)}^{i}\|_{2}-\frac{\|\mathbf{p}_{(t+1)}^{i}\|_{2}^{2}}{2\|\mathbf{p}_{(t)}^{i}\|_{2}})
	\nonumber \\
	&\leq \emph{f}(\mathbf{P}_{(t)})+\lambda_{1} \sum_{i=1}^{N} \frac{\|\mathbf{p}_{(t)}^{i}\|^{2}}{2\|\mathbf{p}_{(t)}^{i}\|}
	+\lambda_{1}\sum_{i=1}^{N}(\|\mathbf{p}_{(t)}^{i}\|_{2}-\frac{\|\mathbf{p}_{(t)}^{i}\|_{2}^{2}}{2\|\mathbf{p}_{(t)}^{i}\|_{2}})
	\end{align}
\end{small}
Thus, we have
\begin{align}\label{Proof5}
&\emph{f}(\mathbf{P}_{(t+1)})+\lambda_{1}\sum_{i=1}^{N}(\|\mathbf{p}_{(t+1)}^{i}\|_{2})
\leq \emph{f}(\mathbf{P}_{(t)})+\lambda_{1}\sum_{i=1}^{N}(\|\mathbf{p}_{(t)}^{i}\|_{2}) \nonumber \\
&\Rightarrow \emph{f}(\mathbf{P}_{(t+1)})+\lambda_{1}\|\mathbf{P}_{(t+1)}\|_{2,1} \leq \emph{f}(\mathbf{P}_{(t)})+ \lambda_{1}\|\mathbf{P}_{(t)}\|_{2,1}
\end{align}
which completes the proof.

\section{Experiments}
\label{Experiments}
In this section, we evaluate GGLS on eight publicly available datasets, and compare with some state-of-the-art methods including traditional
domain adaptation methods and depth domain adaptation methods, and also introduce the parameters analysis and ablation study in this section.
%\begin{figure}
%\centering
%\includegraphics[scale=0.6]{dataset_sample.pdf}
%\caption{Some selected samples from Office+Caltech, USPS, MNIST, ImageNet and VOC2007.}
%\label{dataset_sample}
%\end{figure}

\subsection{Experimental Settings}
Office+Caltech object datasets~\cite{gong2012geodesic} consists of four datasets: Amazon(A), DSLR(D), Webcam(W), and Caltech-256(C) (4 datasets domain adaptation, 4DA). The first three datasets~\cite{Saenko2010Adapting} contains 31 classes, and the last dataset~\cite{griffin2007caltech} contains 256 classes.
There are 10 common classes shared by the four datasets. Furthermore, we consider two types of features: SURF features of 800 dimensions, and Decaf$_{6}$ features of 4096 dimensions~\cite{donahue2014decaf}, which are the activations of the 6th fully connected layer of a convolutional network trained on imageNet. We follow the setting of~\cite{zhang2017joint} to construct 12 cross-domain datasets, which are listed in Tables \ref{Office+Caltech10_SURF},~\ref{Office+Caltech10_DeCaf}, where C$\rightarrow$A represents C is source domain and A is target domain. For the parameters, we set $\beta=0.1$, $\gamma=0.1$, $\lambda_{1}=0.001$, $\lambda_{1}=0.01$, $T = 10$, $d=30$, and $k=3, 9$ for SURF and Decaf$_{6}$ features.

USPS (U)~\cite{lecun1998gradient} and MNIST (M)~\cite{hull1994database} are digit datasets containing handwritten digits from 0-9. There are 10 common classes shared by the two datasets. MNIST dataset contains 60,000 training samples and 10,000 testing samples of size $28\times28$. USPS dataset contains 7,291 training samples and 2,007 testing samples of size $16\times16$. For fair comparison, we extract SURF features of 256 dimensions. We construct two cross-domain datasets: U$\rightarrow$M and M$\rightarrow$U. For the parameters, we set $\beta=0.1$, $\gamma=0.1$, $\lambda_{1}=0.0001$, $\lambda_{1}=0.01$, $T = 10$, $d=30$, and $k=3$.

ImageNet (I) and VOC2007 (V) are large-scale image datasets. We follow the~\cite{fang2013unbiased} to select five common classes shared by the two datasets: bird, cat, chair, dog and person.  We extract Decaf$_{6}$ features of 4096 dimensions and construct two cross-domain datasets: I$\rightarrow$V and V$\rightarrow$I. For the parameters, we set $\beta=0.1$, $\gamma=0.1$, $\lambda_{1}=0.001$, $\lambda_{1}=0.01$, $T = 10$, $d=20$, and $k=5$.

{
CIFAR10(CI) and STL10(S) are both 10 class datasets. CIFAR10 contains 50,000 training samples of size $32\times32$, while STL10 contains 5,000 training samples of size $96\times96$. There are 9 overlapping classes between these two datasets. Following~\cite{roy2019unsupervised} we remove non-overlapping classes - “frog” and “monkey” from these two datasets. We use ResNet50 to extract features and construct two cross-domain datasets: CI$\rightarrow$S and S$\rightarrow$CI. For the parameters, we set $\beta=0.1$, $\gamma=0.1$, $\lambda_{1}=0.001$, $\lambda_{1}=0.01$, $T = 10$, $d=20$, and $k=5$.}

\begin{figure}[!htb]
	\centering
	\includegraphics[scale=0.65]{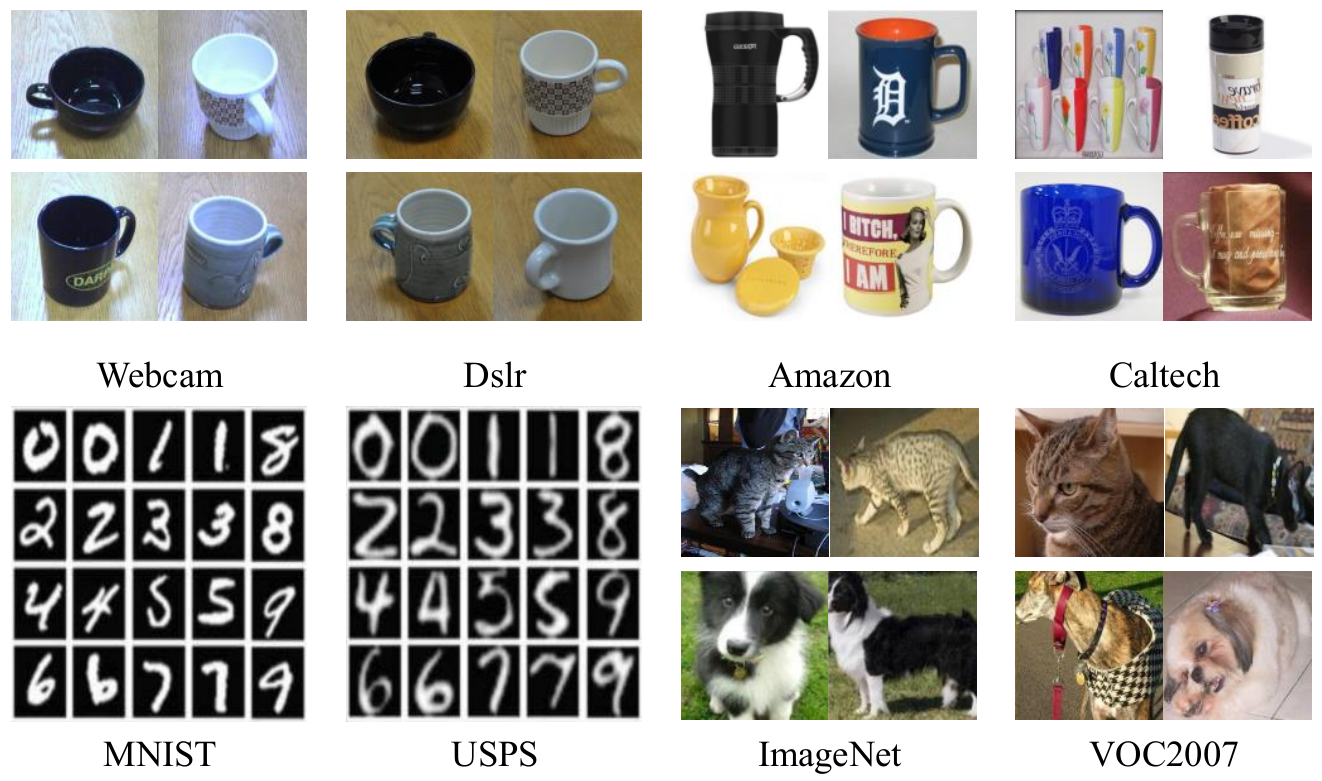}
	\caption{Some selected samples from Office+Caltech, USPS, MNIST, ImageNet and VOC2007.}
	\label{dataset_sample}
\end{figure}

Eight publicly available datasets are shown in Fig.~\ref{dataset_sample}, including Office+Caltech object datasets, which consists of four datasets: Amazon(A), DSLR(D), Webcam(W), and Caltech-256(C), USPS(U), MNIST(M), ImageNet(I) and VOC2007(V).

In this section, the classification accuracy of the target domain is used as the evaluation index, which has been widely applied in the existing methods, namely:
\begin{align}\label{evaluation_index}
Accuracy = \frac{|{{\bf{x}}:{\bf{x}} \in {{\bf{X}}_t} \wedge {{\widehat {\bf{l}}}_t} = {{\bf{l}}_t}}|}{|{{\bf{x}}:{\bf{x}} \in {{\bf{X}}_t}}|}
\end{align}

\subsection{Results}
In this section, the comparison methods include traditional domain adaptation methods and depth domain adaptation methods.

Traditional domain adaptation methods:
\begin{enumerate}
	\item  $\mathbf{kNN}$(k=1) and $\mathbf{PCA}$
	\item  $\mathbf{GFK}$(Geodesic Flow Kernel)~\cite{gong2012geodesic}:performs manifold feature learning
	\item  $\mathbf{TCA}$(Transfer Component Analysis)~\cite{pan2010domain}: performs marginal distribution adaptation
	\item  $\mathbf{JDA}$(Joint Distribution Alignment)~\cite{long2013transfer}: performs marginal and conditional distribution adaptation
	\item {$\mathbf{SA}$(Subspace Alignment)~\cite{fernando2013unsupervised}: performs subspace alignment}
	\item  $\mathbf{\mathbf{TJM}}$(Transfer Joint Matching)~\cite{long2014transfer}:performs marginal distribution adaptation with sample selection of source domain
	\item  $\mathbf{MLS}$(Multiscale Landmarks Selection)~\cite{aljundi2015landmarks}: performs landmark selection on both domains
	\item  $\mathbf{\mathbf{SCA}}$(Scatter Component Analysis)~\cite{ghifary2017scatter}: performs scatter adaptation in subspace
	\item  $\mathbf{JGSA}$(Joint Geometrical and Statistical Alignment)~\cite{zhang2017joint}:performs marginal and conditional distribution adaptation with label information
	\item  $\mathbf{TIT}$(Transfer Independently Together)~\cite{li2018transfer}: performs landmark selection based on graph optimization
	\item  $\mathbf{MEDA}$(Manifold Embedded Distribution Alignment)~\cite{wang2018visual}: performs distribution alignment based on structural risk minimization
	\item  {$\mathbf{CAPLS}$( Confidence-Aware Pseudo Label Selection)~\cite{wang2019unifying}: performs confidence-aware pseudo label selection scheme}
	\item { $\mathbf{SPL}$( Selective Pseudo-Labeling)~\cite{wang2020unsupervised}: performs selective pseudo-labeling strategy based on structured prediction}
	
\end{enumerate}

Deep domain adaptation approaches:
\begin{enumerate}
	\item $\mathbf{AlexNet}$~\cite{krizhevsky2012imagenet}
	\item $\mathbf{DDC}$(Deep Domain Confusion)~\cite{tzeng2014deep}:performs single-layer deep adaptation with MMD loss
	\item $\mathbf{DAN}$(Deep Adaptation Network)~\cite{long2014domain} performs multi-layer adaptation with multiple kernel MMD
	\item $\mathbf{DCORAL}$(Deep CORAL)~\cite{sun2016deep}:performs neural network with CORAL loss
	\item $\mathbf{DUCDA}$(Deep Unsupervised Convolutional Domain Adaptation)~\cite{zhuo2017deep}: performs adaptation with attention and CORAL loss
	\item {$\mathbf{ReverseGrad}$(Reverse Gradient)~\cite{ganin2015unsupervised}: performs gradient reversal layer}
	\item {$\mathbf{DRCN}$(Deep Reconstruction Classification Network )~\cite{ghifary2016deep}: performs shared encoding representation for alternately learning (source) label prediction and (target) data reconstruction}
	\item {$\mathbf{AutoDIAL}$(Automatic DomaIn Alignment Layers)~\cite{maria2017autodial}: performs domain alignment layers}
	\item {$\mathbf{DWT}$(Domain-specific Whitening Transform)~\cite{roy2019unsupervised}: performs domain specific feature whitening}
	
\end{enumerate}

\begin{table*}[!htb]
	\caption{Accuracy results (\%) on Office+Caltech10 datasets using SURF features.}
	\centering
	\resizebox{!}{22mm}
	{
		\begin{tabular}{ccccccccccccc}
			\hline
			Data               &1NN    & PCA   & GFK   &  TCA  &  JDA   & TJM   & SCA        & MLS     & JGSA  & TIT    &  MEDA & GGLS \\
			\hline
			C$\rightarrow$ A     & 23.7  & 39.5  & 46.0  & 45.6  &  43.1  &46.8    &45.6 &   53.6        &53.1  &  59.7  &56.5  & $\mathbf{58.2}$ \\
			C$\rightarrow$ W     & 25.8  & 34.6  & 37.0  & 39.3  & 39.3   &39.0    &40.0 &    44.1       &48.5  &  51.5  &53.9  & $\mathbf{55.3} $ \\
			C$\rightarrow$ D     & 25.5  & 44.6  & 40.8  &45.9   & 49.0   &44.6    &47.1 &     45.2      &48.4  &  48.4  &50.3  & $ \mathbf{51.6}$ \\
			A$\rightarrow$ C     & 26.0  &39.0   &40.7   & 42.0  & 40.9   &39.5    &39.7 &    45.0      &41.5  &  47.5  &43.9  &  $\mathbf{48.6}$ \\
			A$\rightarrow$ W     & 29.8  & 35.9  &37.0   &40.0   &38.0    &42.0    &34.9 &     41.1     &45.1  &  45.4  &$\mathbf{53.2}$ &  51.2  \\
			A$\rightarrow$ D     & 25.5  & 33.8  &40.1   &35.7   &42.0    &45.2    &39.5  &    39.5        &45.2  &  47.1  &45.9  & $ \mathbf{51.0}$\\
			W$\rightarrow$ C     & 19.9  & 28.2  &24.8   &31.5   &33.0    &30.2    &31.1 &     31.6          &33.6  &$\mathbf{34.9 }$   & 34.0&  32.6  \\
			W$\rightarrow$ A     & 23.0  & 29.1 &27.6    &30.5   & 29.8   &30.0    &30.0 &      35.9        &40.8  &40.2    &$\mathbf{42.7} $&  40.7  \\
			W$\rightarrow$ D     & 59.2  &89.2  & 85.4   &91.1   &$\mathbf{92.4}$   &89.2 &87.3&    84.7      &88.5  &87.9    &88.5& 90.4\\
			D$\rightarrow$ C     & 26.3  &29.7  & 29.3   &33.0   & 31.2   & 31.4   & 30.7 &      39.2        &30.3 &$\mathbf{36.7 }$   &34.9 & 35.2 \\
			D$\rightarrow$ A     & 28.5  &33.2 &28.7     &32.8   & 33.4   &32.8    &31.6&      34.6         &38.7  &42.1    &41.2 & $ \mathbf{44.3}$\\
			D$\rightarrow$ W     & 63.4  &86.1 &80.3     &87.5   &89.2    &85.4    &84.4 &    82.4      &$\mathbf{93.2}$  &84.8    &87.5 & 90.2\\
			Average             & 31.4  &43.6 &43.1     &46.2   &46.8    &46.3    &45.2  &  48.1         &50.6 &52.2& 52.7   &$\mathbf{54.1}$\\
			\hline
		\end{tabular}
	}
	\label{Office+Caltech10_SURF}
\end{table*}

\begin{table*}[!htb]
	\caption{{Comprisons with traditional methods on Office+Caltech10 datasets using DeCaf$_{6}$ features.}}
	\centering
	\resizebox{!}{22mm}
	{
		\begin{tabular}{ccccccccccccc}
			\hline
			Data                 &1NN    & PCA   & GFK   &  TCA  &  JDA   & TJM   & SCA&  JGSA & {CAPLS}& MEDA & {SPL} &GGLS \\
			\hline
			C$\rightarrow$ A     & 87.3  &88.1   &88.2    &89.8  &89.6    &88.8    &89.5 &91.1 &90.8&92.3&  92.7& $\mathbf{94.9}$\\
			C$\rightarrow$ W     & 72.5  &83.4  &77.6   &78.3    &85.1    &81.4    &85.4 &86.8 &85.4&91.2& 93.2&$\mathbf{93.6}$\\
			C$\rightarrow$ D     & 79.6  &84.1  &86.6    &85.4   &89.8    &84.7    &87.9 &93.6 &$\mathbf{95.5}$ &89.8& 98.7&94.9\\
			A$\rightarrow$ C     & 71.7 &79.3   &79.2   &82.6    &83.6    &84.3    &78.8 &84.9 &86.1&89.2&  87.4&$\mathbf{89.7}$\\
			A$\rightarrow$ W     & 68.1 &70.9   &70.9    &74.2   &78.3    &71.9    &75.9 &81.0&87.1 &$\mathbf{90.8}$&  95.3 &87.1\\
			A$\rightarrow$ D     & 74.5  &82.2  &82.2    &81.5   &80.3    &76.4    &85.4 &88.5&$\mathbf{94.9}$ &87.9&  89.2& 88.5\\
			W$\rightarrow$ C     & 55.3  &70.3  &69.8    &80.4   &84.8    &83.0    &74.8 &85.0&88.2 &88.7& 87.0 & $\mathbf{89.9}$\\
			W$\rightarrow$ A     & 62.6 &73.5   & 76.8   &84.1   &90.3    &87.6    &86.1 &90.7&92.3 &92.5&  92.0& $\mathbf{94.2}$\\
			W$\rightarrow$ D     & 98.1 & 99.4  &$\mathbf{100}$   &$\mathbf{100}$  &$\mathbf{100}$   &$\mathbf{100}$   &$\mathbf{100}$ &$\mathbf{100}$&$\mathbf{100}$&99.4&$\mathbf{100}$ & $\mathbf{100}$\\
			D$\rightarrow$ C     & 42.1 &71.7   & 71.4   &82.3   & 85.5   & 83.8   & 78.1 &86.2&88.8 &88.2&88.6&$\mathbf{90.2}$\\
			D$\rightarrow$ A     & 50.0 &79.2   &76.3   &  89.1   & 91.7   &90.3    &90.0 &92.0&93.0 &92.6& 92.9& $\mathbf{94.1}$\\
			D$\rightarrow$ W     & 91.5 & 98.0   &99.3  &  99.7  & 99.7    & 99.3   &98.6 & 99.7&$\mathbf{100}$ & 98.6& 98.6&$\mathbf{100}$\\
			Average             & 71.1   & 81.7  & 81.5  &85.6   &88.2    &86.0    &85.9  &90.0&91.8 &91.8& 93.0&$\mathbf{93.1}$\\
			\hline
		\end{tabular}
	}
	\label{Office+Caltech10_DeCaf}
\end{table*}

\begin{table*}[!htb]
	\caption{Comprisons with deep learning based methods on Office+Caltech10 datasets using DeCaf6 features.}
	\centering
	\resizebox{!}{25mm}
	{
		\begin{tabular}{lrrrrrr}
			\hline
			Data                 & AlexNet	& DDC	& DAN& 	DCORAL	& DUCDA	& GGLS \\
			\hline
			C$\rightarrow$ A     & 91.9	&91.9	&92.0	&92.4	&92.8	& $\mathbf{94.9}$\\
			C$\rightarrow$ W     &83.7&	85.4	&90.6	&91.1	&91.6& $\mathbf{93.6}$\\
			C$\rightarrow$ D     & 87.1&	88.8&	89.3	&91.4	&91.7& $\mathbf{94.9}$\\
			A$\rightarrow$ C     & 83.0	&85.0	&84.1	&84.7	&84.8&  $\mathbf{89.7}$\\
			A$\rightarrow$ W     & 79.5	&86.1	&$\mathbf{91.8}$	& $-$	& $-$&  87.1\\
			A$\rightarrow$ D     & 87.4	&89.0&	$\mathbf{91.7}$&	$-$&$-$&$88.5$\\
			W$\rightarrow$ C     & 73.0&	78.0	&81.2	&79.3	&80.2& $\mathbf{89.9}$\\
			W$\rightarrow$ A     & 83.8	&84.9	&92.1&		$-$&$-$&  $\mathbf{94.2}$\\
			W$\rightarrow$ D     & $\mathbf{100}$	&$\mathbf{100}$	&$\mathbf{100}$&	$-$	& $-$& $\mathbf{100}$\\
			D$\rightarrow$ C     & 79.0&	81.1	&80.3	&82.8	&82.5&$\mathbf{90.2}$\\
			D$\rightarrow$ A     & 87.1	&89.5	&90.0	&	$-$&$-$&  $\mathbf{94.1}$\\
			D$\rightarrow$ W     & 97.7	&98.2&	98.5&	$-$	&$-$& $\mathbf{100}$\\
			Average             &86.1	&88.2	&90.1&	$-$	&$-$& $\mathbf{93.1}$\\
			\hline
		\end{tabular}
	}
	\label{Office+Caltech10_DeCaf_deep}
\end{table*}

\begin{table*}[!htb]
	\caption{Accuracy results(\%) on USPS+MNIST and ImageNet+VOC2007 datasets.}
	\centering
	{
		\begin{tabular}{lrrrrrrrrrrr}
			\hline
			Data                 &1NN    & PCA   & GFK   &  TCA  &  JDA   & TJM   & SCA&  JGSA &  MEDA & GGLS \\
			\hline
			U$\rightarrow$ M     &44.7   &45.0    &46.5  &51.2   & 59.7   &52.3   &48.0  &68.2 &72.1& $\mathbf{74.7}$ \\
			M$\rightarrow$ U     & 65.9 &66.2     &61.2   &56.3  &67.3    &63.3   &65.1  &80.4 &89.5& $\mathbf{90.6}$\\
			I$\rightarrow$ V     & 50.8 & 58.4    &59.5   & 63.7 &63.4     &63.7  &$-$     & 52.3&67.3 &$\mathbf{68.1}$\\
			V$\rightarrow$ I     & 38.2 &65.1     &73.8   &64.9  &70.2    &73.0  &$-$      &70.6  &74.7&$\mathbf{78.3}$\\
			Average              & 49.9 &58.7     &60.2    &59.0 &65.1    &63.1   &$-$     &67.9  &75.9 &$\mathbf{77.9}$\\
			\hline
		\end{tabular}
	}
	\label{USPS+MNIST}
\end{table*}
\begin{table*}[!htb]
	\caption{{Accuracy results(\%) on CIFAR10 and STL10 datasets.}}
	\centering
	{
		\begin{tabular}{cccccccccc}
			\hline
			Data                   & SA    &  ReverseGrad  &  DRCN   & AutoDIAL   & DWT&  GGLS \\
			\hline
			CI$\rightarrow$ S       &62.9     &66.1   & 66.4   &79.1   &79.7  & $\mathbf{80.1}$ \\
			S$\rightarrow$ CI       &54.0     &56.9   &58.9    &70.2   &$\mathbf{71.2}$  & 70.6\\
			Average               &58.5     &61.5   &62.7   &74.6   &$\mathbf{75.5}$  & 75.4\\
			\hline
		\end{tabular}
	}
	\label{CIFAR+STL}
\end{table*}

\begin{table*}
	\caption{{Results of Wilcoxon rank sum tests for GGLS}}
	\label{Wilcoxon}
	\centering
	\resizebox{!}{22mm}
	{
		\begin{tabular}{cc||cc||cc||cc}
			\hline
			\multicolumn{2}{c||}{Results of Tables\ref{Office+Caltech10_SURF} based} &\multicolumn{2}{c||}{Results of Tables~\ref{Office+Caltech10_DeCaf} and ~\ref{Office+Caltech10_DeCaf_deep} based}&\multicolumn{2}{c||}{Results of Tables~\ref{USPS+MNIST} based}&\multicolumn{2}{c}{Results of Tables~\ref{CIFAR+STL} based}\\
			\hline
			GGLS  vs. &p-values & GGLS  vs. &	p-values& GGLS  vs. &	p-values& GGLS  vs. &	p-values\\
			\hline
			1NN  & 0.003 &1NN & 0.001 &1NN  & 0.009 &SA& 0.121\\
			PCA  & 0.024& PCA&0.003& PCA & 0.009&  ReverseGrad  & 0.121\\
			GFK  &0.035 & GFK& 0.004& GFK & 0.016&  DRCN & 0.121 \\
			TCA & 0.073&  TCA& 0.008 &  TCA&0.009& AutoDIAL &0.439 \\
			JDA   &0.083  &  JDA &0.038  &  JDA & 0.016 & DWT&1.000 \\
			TJM  & 0.065& TJM &0.013 & TJM&0.016&&\\
			SCA&0.033& SCA& 0.011&   JGSA& 0.175&&\\
			MLS &0.184&  JGSA& 0.094&  MEDA& 0.403 &&\\
			JGSA&0.326& CAPLS&0.470 && &&\\
			TIT&0.525& MEDA&  0.299 && && \\
			MEDA &0.686& SPL& 0.729 && &&\\
			&&AlexNet&0.011 && &&\\
			&& DDC& 0.021 && &&\\  
			&&DAN& 0.248 && &&\\
			\hline
		\end{tabular}
	}	
\end{table*}

{Tables~\ref{Office+Caltech10_SURF},~\ref{Office+Caltech10_DeCaf},~\ref{Office+Caltech10_DeCaf_deep},~\ref{USPS+MNIST} and~\ref{CIFAR+STL} show the accuracy results on the different datasets.} From Tables~\ref{Office+Caltech10_SURF},~\ref{Office+Caltech10_DeCaf} and \ref{USPS+MNIST}, it can be observed  that transfer learning methods outperform 1NN and PCA, which indicates transfer learning is valuable and practical in real world applications for classification.
GGLS outperforms the state-of-the-art methods in 19 out of 28 cases, and the average accuracy is 74.2\%. Besides, GGLS achieves the best accuracy in 24 out of 28 cases compared with the baseline method MEDA. These results indicate that GGLS can deal with distribution divergence better. Since these datasets have a large differences, it demonstrates that GGLS has strong generalization ability and robustness in domain adaptation.
{From Table~\ref{Office+Caltech10_DeCaf_deep} and~\ref{CIFAR+STL}, we can see that GGLS also outperforms most of the deep learning based methods, and only worse than DWT, which use deep learning technology to construct domain-specific alignment layers. This layer projects the source domain and target domain into a common spherical distribution. However, DWT often tunes a lot of hyperparameters before obtaining the optimal results. Compared with DWT, GGLS has a more simple architecture and fewer parameters, which can be set easily. }
Therefore, GGLS  has high accuracy and efficiency.
Furthermore, we pay attention to the performance of MLS, TJM, and GGLS, which use samples reweighting. We find that GGLS outperforms the other three methods in many cases. It indicates that it is better to select the landmarks of different samples independently, rather than simply select the same landmarks of samples.
GFK and MEDA learn feature in the manifold, and JGSA preserves discriminative information from the source domain, yet none of them outperforms GGLS in average. It exactly demonstrates that GGLS with optimizing them together is more optimal than optimizing them separately. MEDA learns feature in manifold and preserves discriminative information from the source domain, but it does not consider the samples reweighting. As a result, its performance is worse than GGLS.  {Compared with TIT, which also combine sample reweighting and feature matching, GGLS has better performance. This shows that our combination method is more effective.}

{Based on the results of Tables~\ref{Office+Caltech10_SURF},~\ref{Office+Caltech10_DeCaf},~\ref{Office+Caltech10_DeCaf_deep},~\ref{USPS+MNIST} and~\ref{CIFAR+STL} , we further conduct a Wilcoxon rank sum statistic for GGLS with other methods shown in Table~\ref{Wilcoxon}. The Wilcoxon rank sum test tests the null hypothesis that two sets of measurements are drawn from the same distribution. The alternative hypothesis is that values in one sample are more likely to be larger than the values in the other sample.
The p-value denotes the probability of observing the given result by chance if the null hypothesis is true. Small values of p cast
doubt on the validity of the null hypothesis. That is to say, the greater the value of p is, the smaller the significant difference between the two classes is; and vice versa. As we can see from Table~\ref{Wilcoxon},
there are significantly difference (p-value below 0.1) between GGLS and many compared methods. In other words, GGLS is significantly better than these methods. For other methods like JSGA, TIT, MEDA, CAPLS, although the difference is not significant, the p-value is also small. From the results of Tables~\ref{CIFAR+STL} based in Table~\ref{Wilcoxon}, we can observe that the p-value of GGLS and DWT is 1.000, which further demonstrated that our method is comparable to DWT.}

\begin{figure*}[t]
	\centering
	\subfigure[]{\includegraphics[width=1.8in]{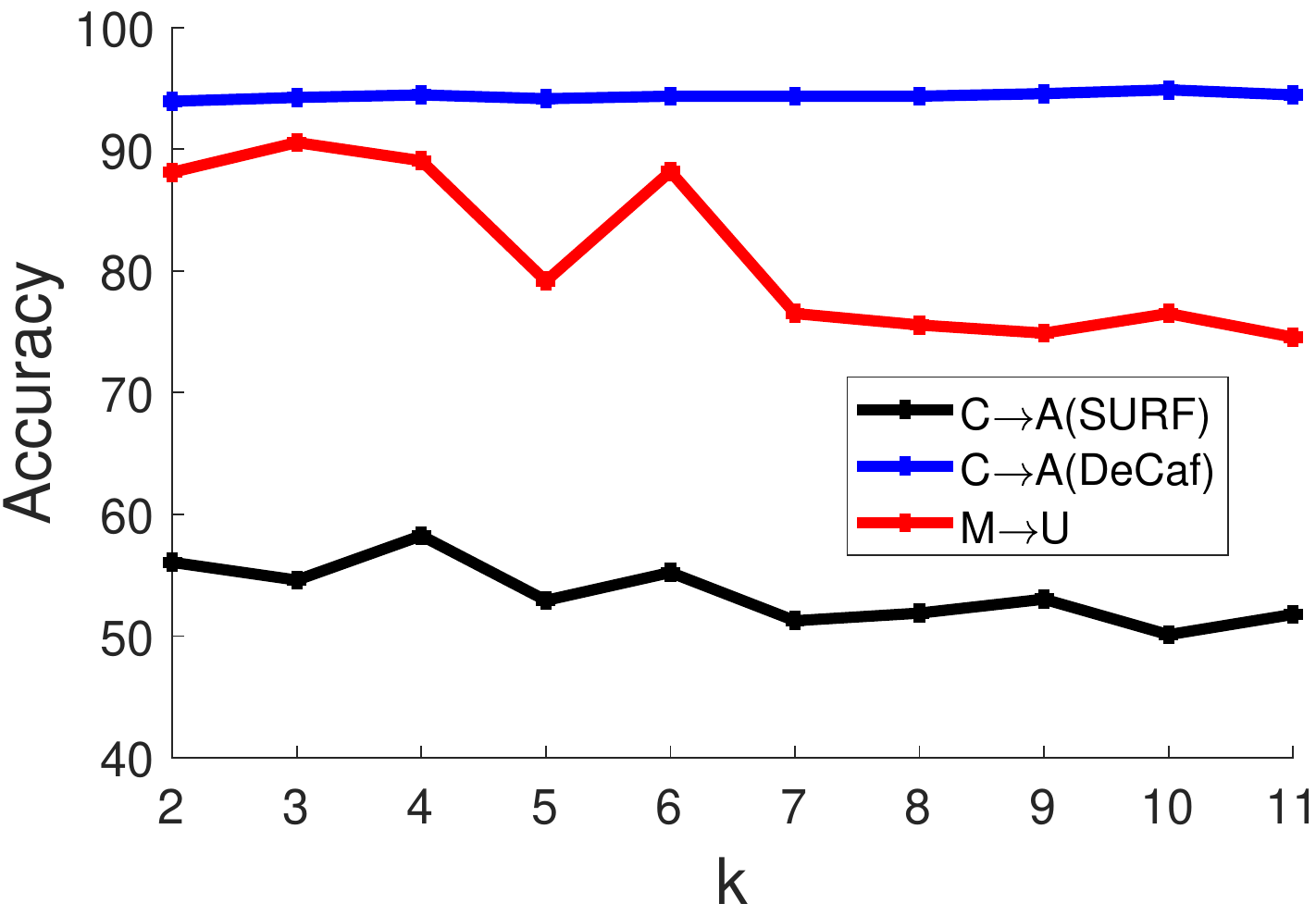}\label{Neighobr}}
	\subfigure[]{\includegraphics[width=1.8in]{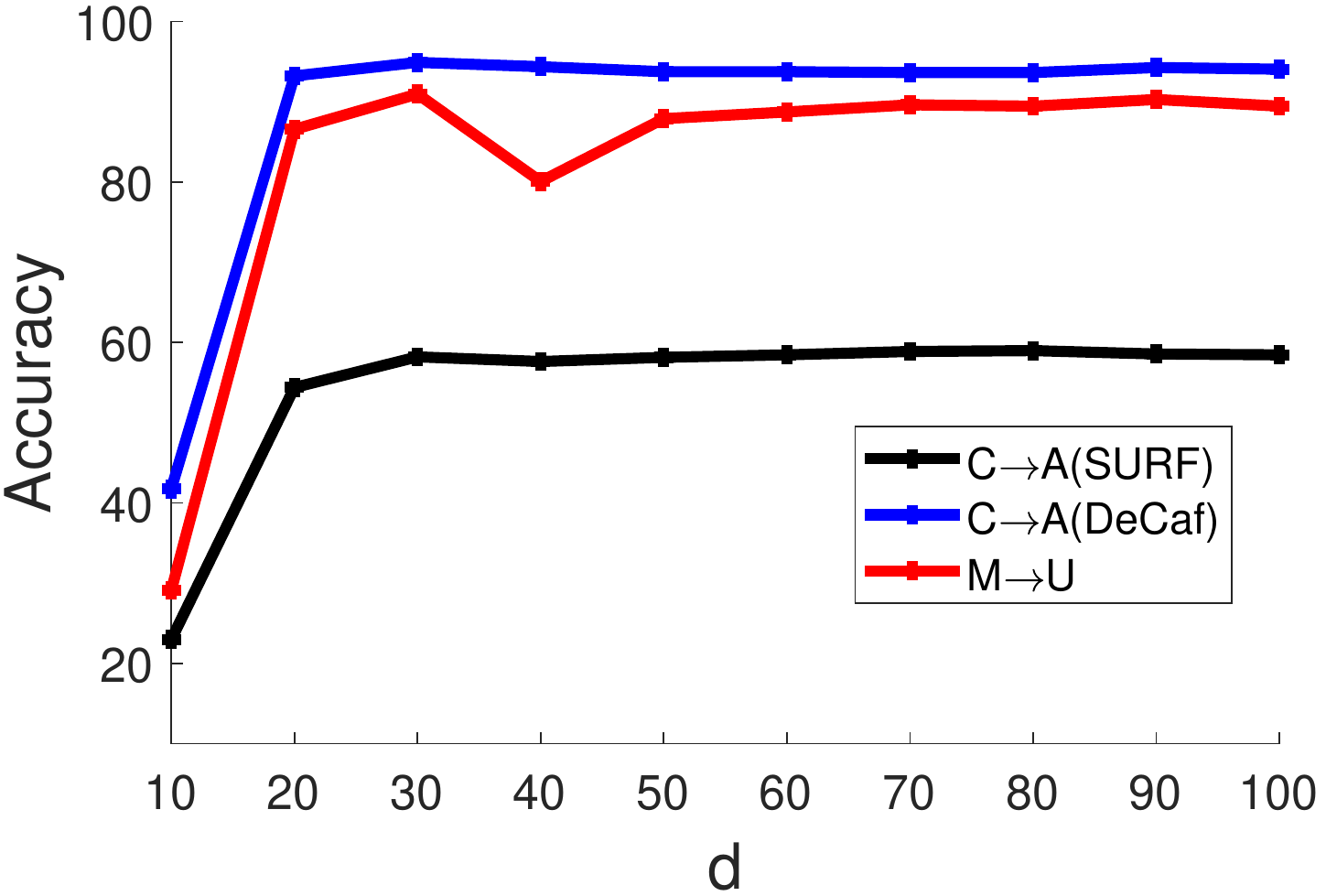}\label{Subspace_dimension}}
	\subfigure[]{\includegraphics[width=1.7in]{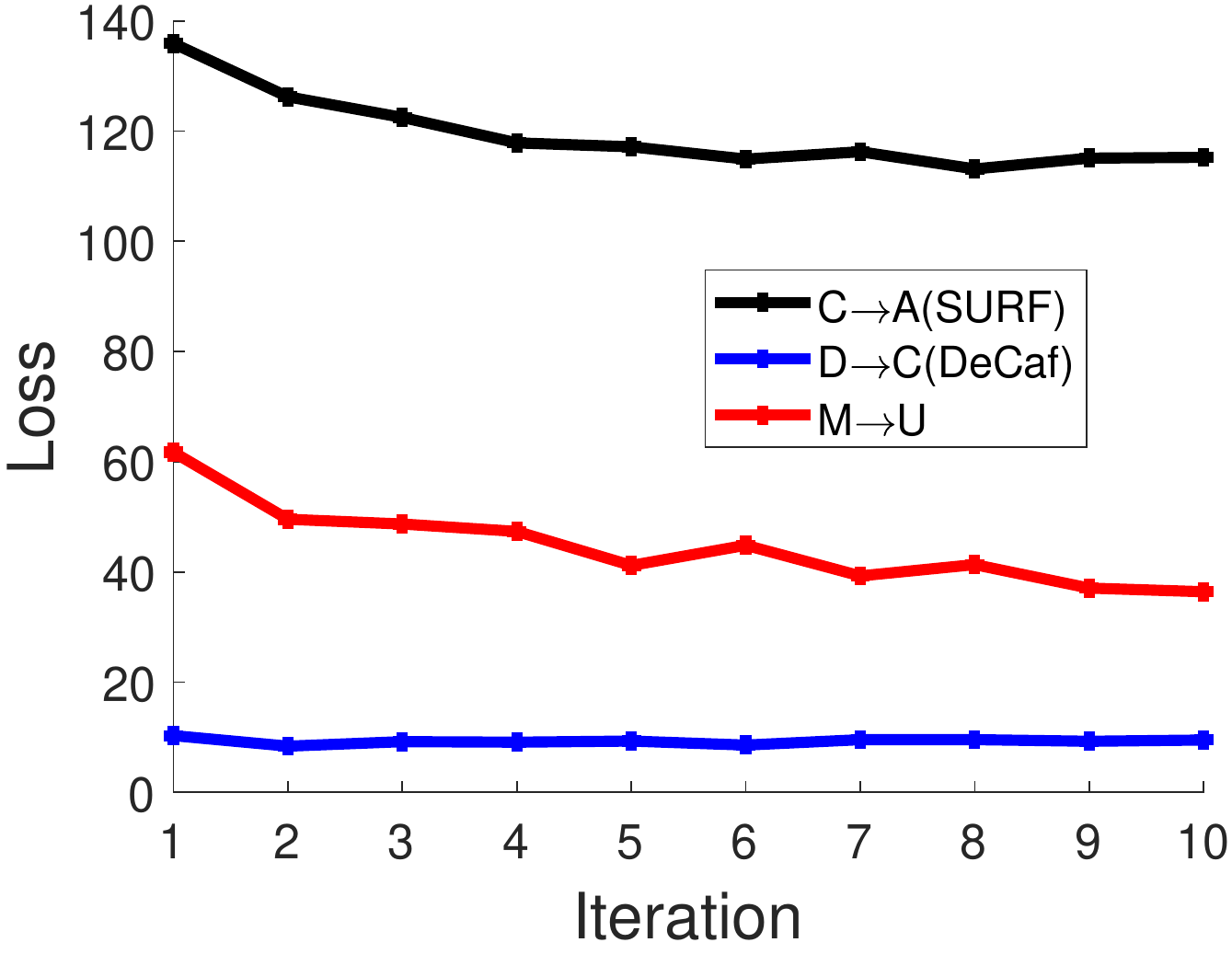}\label{Iteration}}
	\caption{Domain adaptation results. (a) shows the results of different neighbor numbers; (b) shows the results of different subspace dimension. (c) shows the convergence curves.}
	\label{neighbor_dimensionality_convergence}
\end{figure*}

\begin{figure*}[!htb]
	\centering
	\subfigure[]{\includegraphics[width=1.8in]{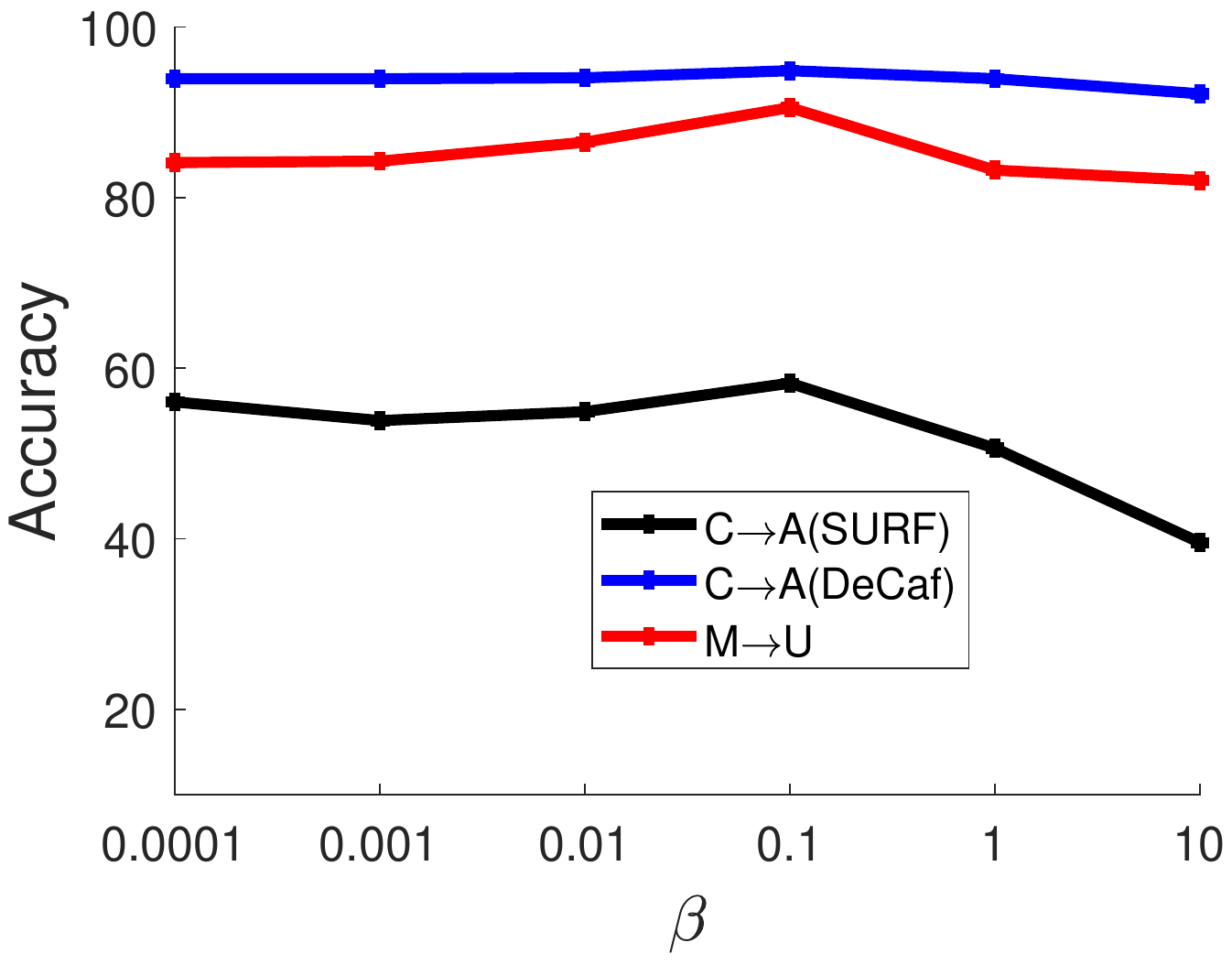}\label{Beta}}
	\subfigure[]{\includegraphics[width=1.8in]{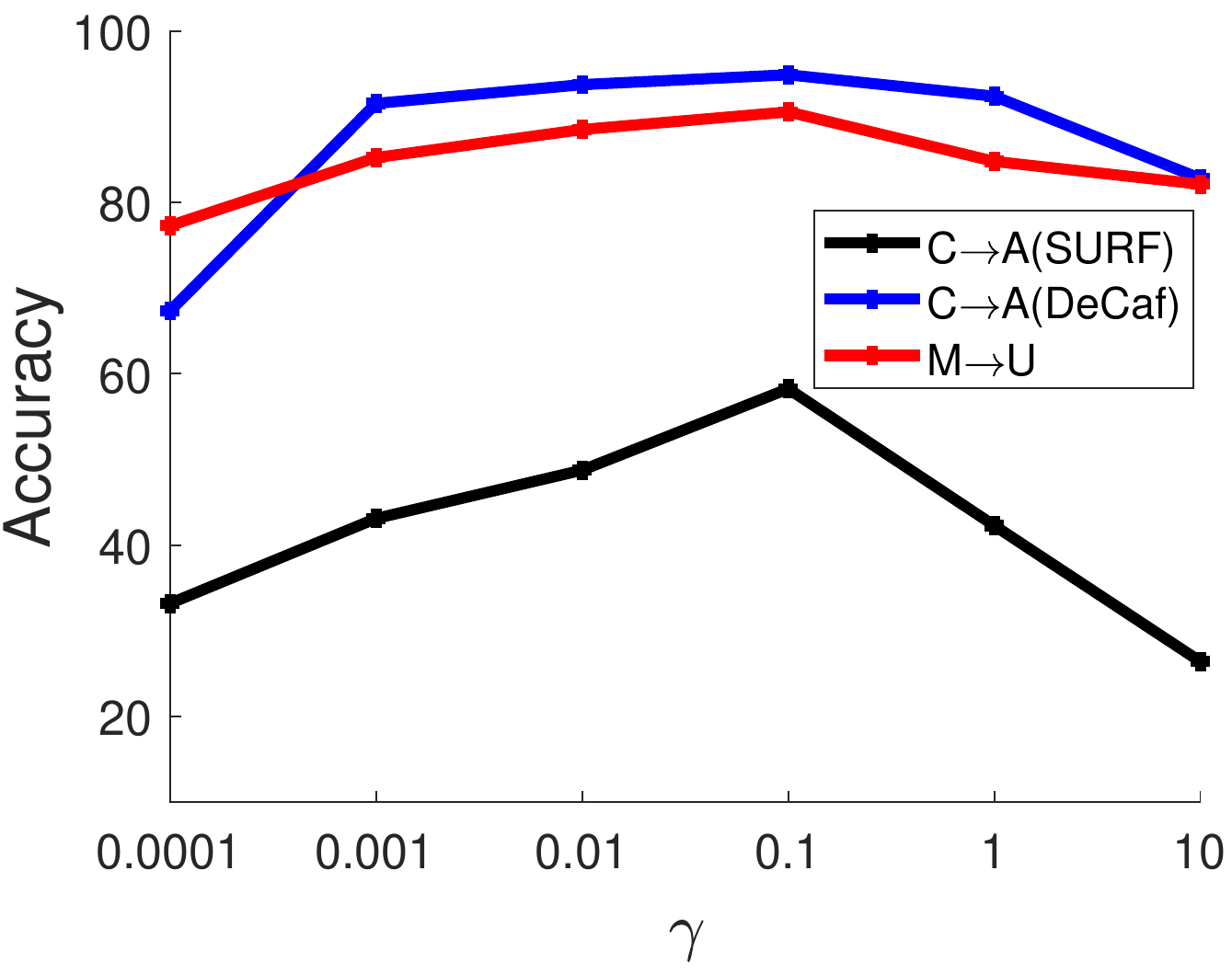}\label{Gamma}}
	\subfigure[]{\includegraphics[width=1.8in]{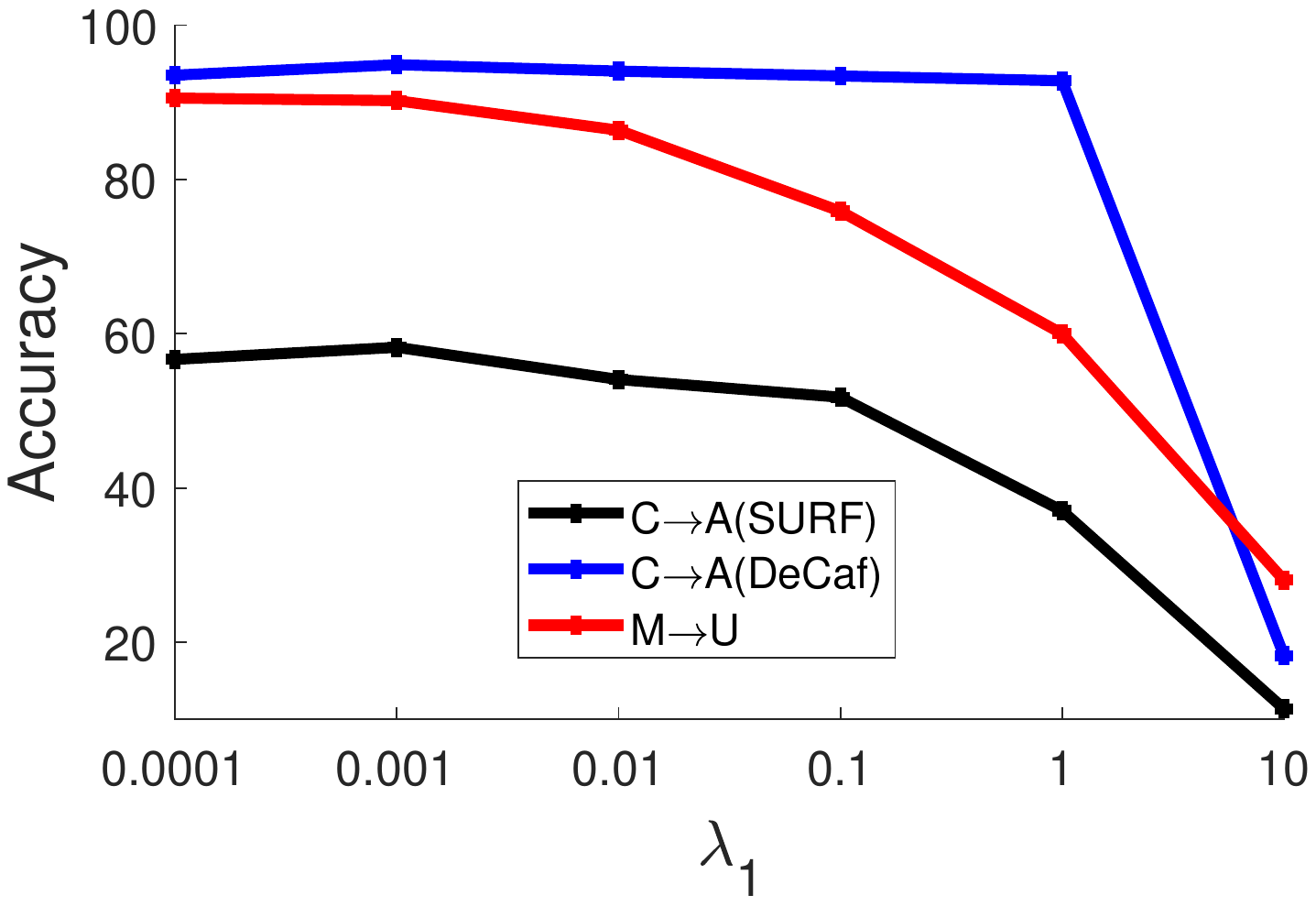}\label{Lambda1}}
	\subfigure[]{\includegraphics[width=1.8in]{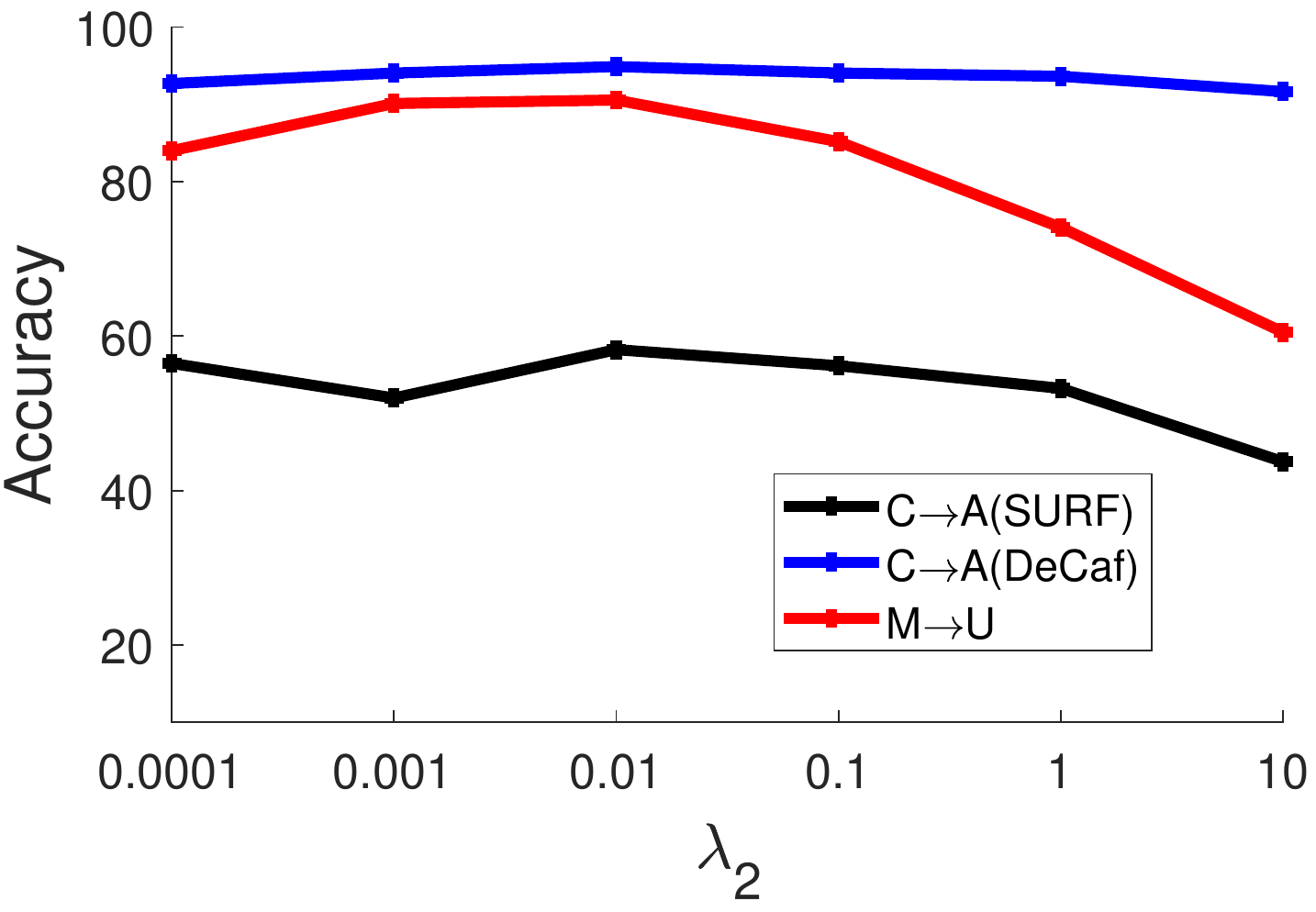}\label{Lambda2}}
	\caption{Accuracy w.r.t different $\beta$, $\gamma$, $\lambda_1$, $\lambda_2$, respectively.}
	\label{parameter_values}
\end{figure*}
\subsection{Parameter Analysis}
We show further details for parameter selection including the number of neighbors $k$, subspace dimension $d$, iteration number $T$, and other regularization parameters $\beta, \gamma, \lambda_{1}, \lambda_{2}$. Fig.~\ref{Neighobr}-\ref{Iteration} show the effects of $k$, $d$ and $T$, respectively. We can observe that GGLS reaches the better performance when $k$ takes $3, 9, 3$ for $\text{C}\rightarrow A$(SURF), $C\rightarrow A$(DeCaf) and $M\rightarrow U$, respectively, and shows a high robustness  for $d$ from $[20, 100]$. For the number of iteration $T$, the results can be converged to the optimum value after a few iterations. Therefore, we set $T=10$. Fig.~\ref{Beta}-\ref{Lambda2} show the effects of regularization parameters $\beta$, $\gamma$, $\lambda_{1}$, $\lambda_{2}$, respectively. We find that GGLS can achieve a robust performance in a wide range of parameter values, whose optimal ranges are $\beta\in[0.0001,1]$, $\gamma\in[0.001,1]$, $\lambda_{1}\in[0.0001,0.1]$ and $\lambda_{2}\in[0.0001,1]$, respectively. When these values of parameters are too large, the learned transformation matrix may be overfitting and thus the performance degrades.

\subsection{Ablation Study}
\begin{figure}
	\centering
	\includegraphics[scale=0.5]{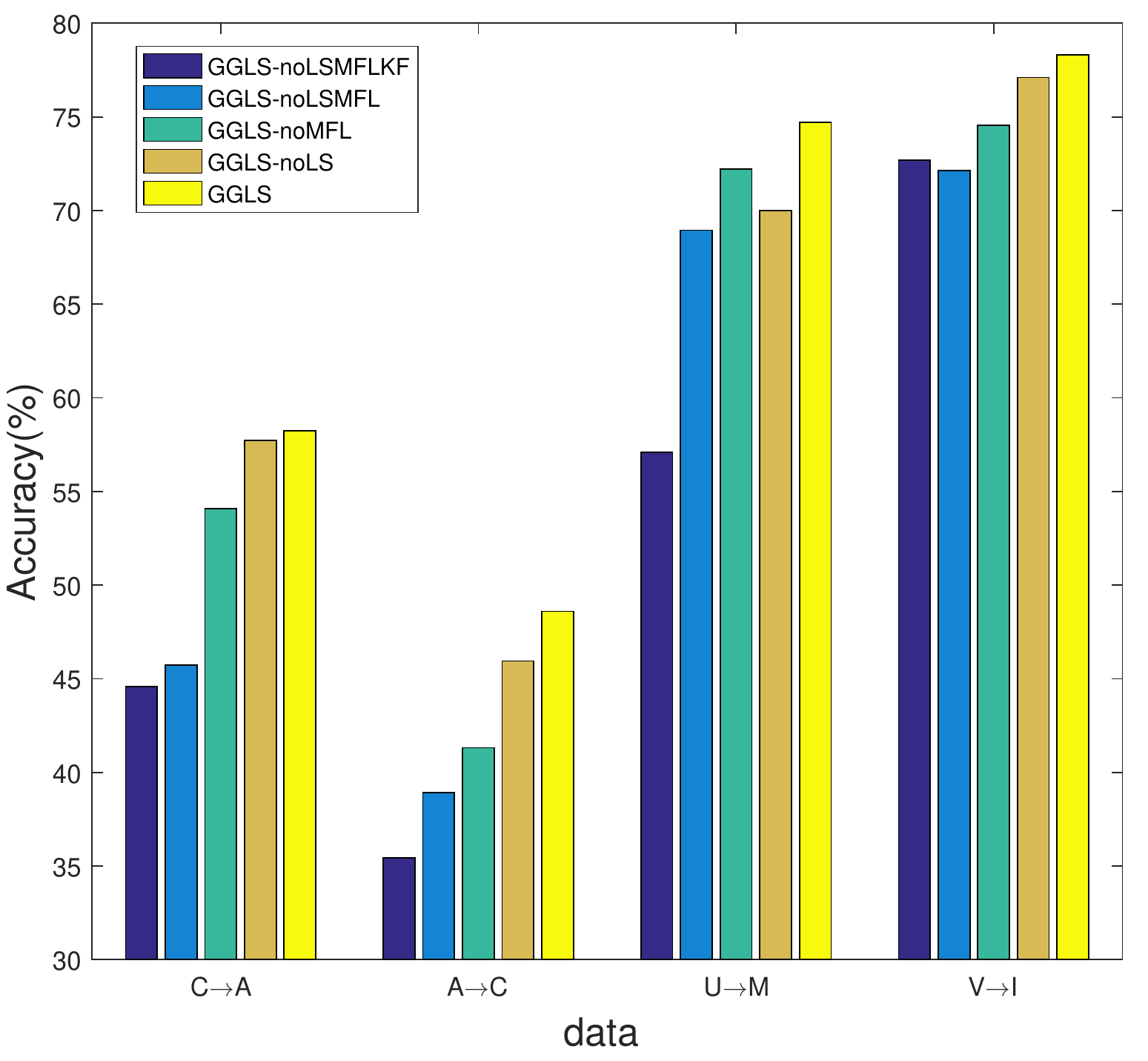}
	\caption{Evaluation of landmark selection and manifold feature learning.}
	\label{Nolandmark_NOmanifold}
\end{figure}
\begin{figure*}[!htb]
	\centering
	\subfigure[Decaf]{\includegraphics[width=1.5in]{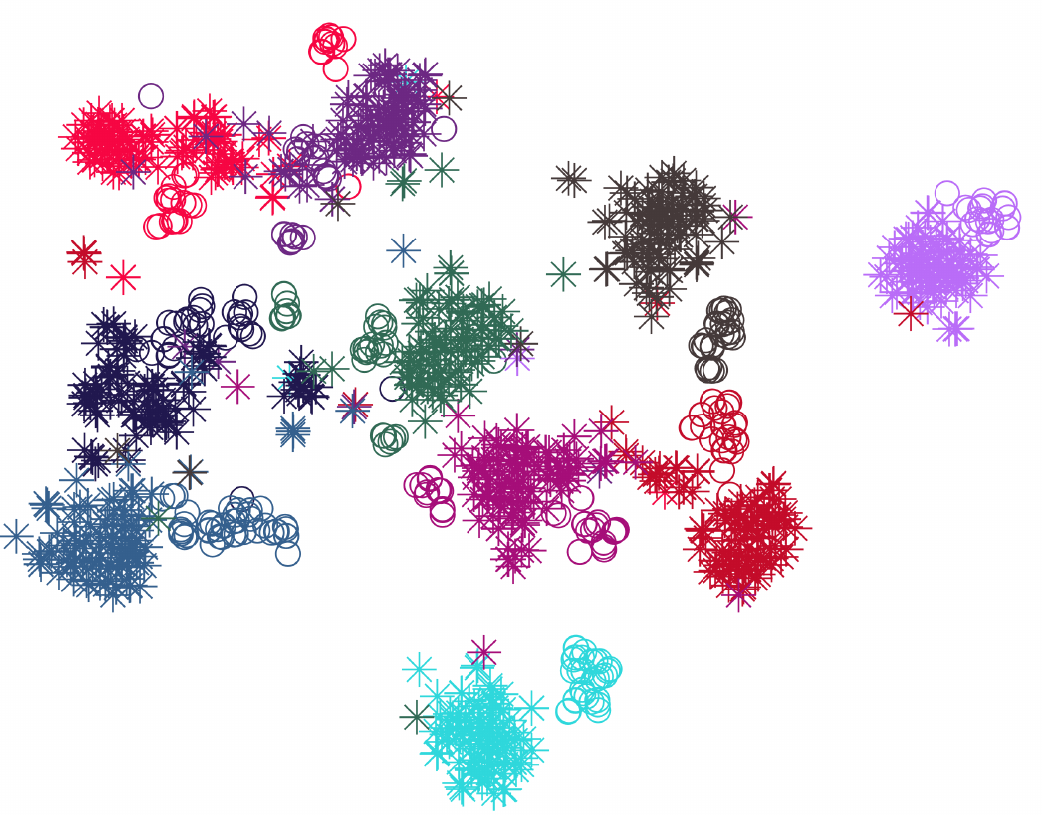}\label{AW1}}
	\subfigure[GGLS-noLS]{\includegraphics[width=1.5in]{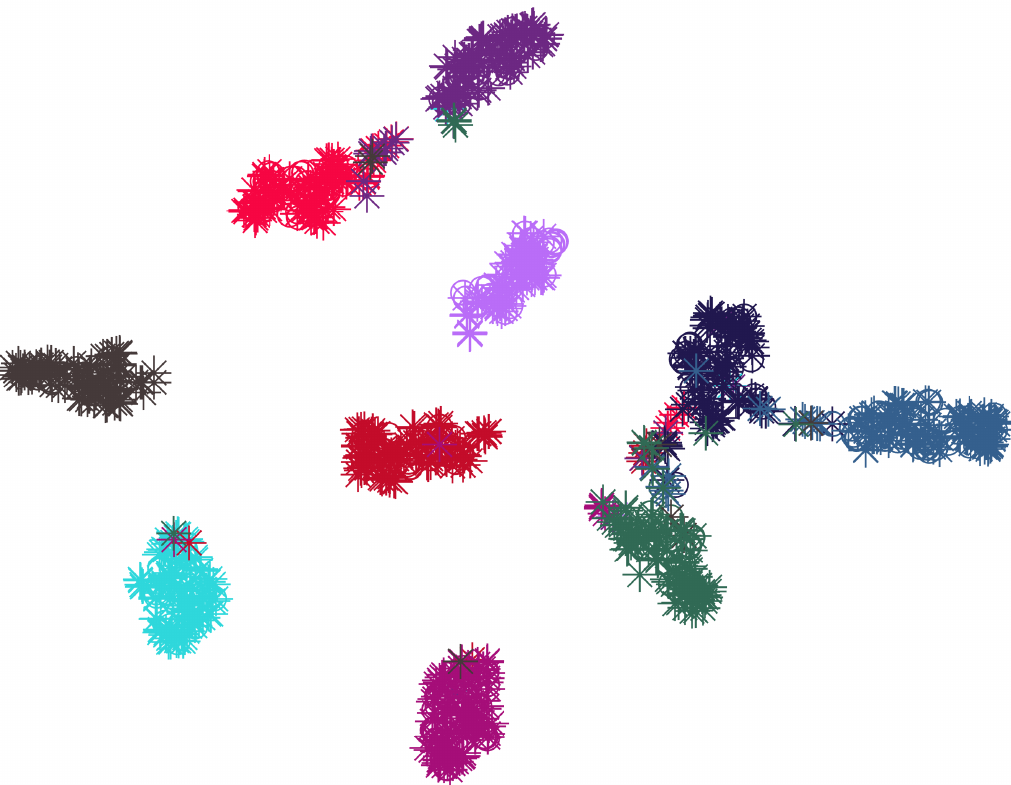}\label{AW22}}
	\subfigure[GGLS]{\includegraphics[width=1.5in]{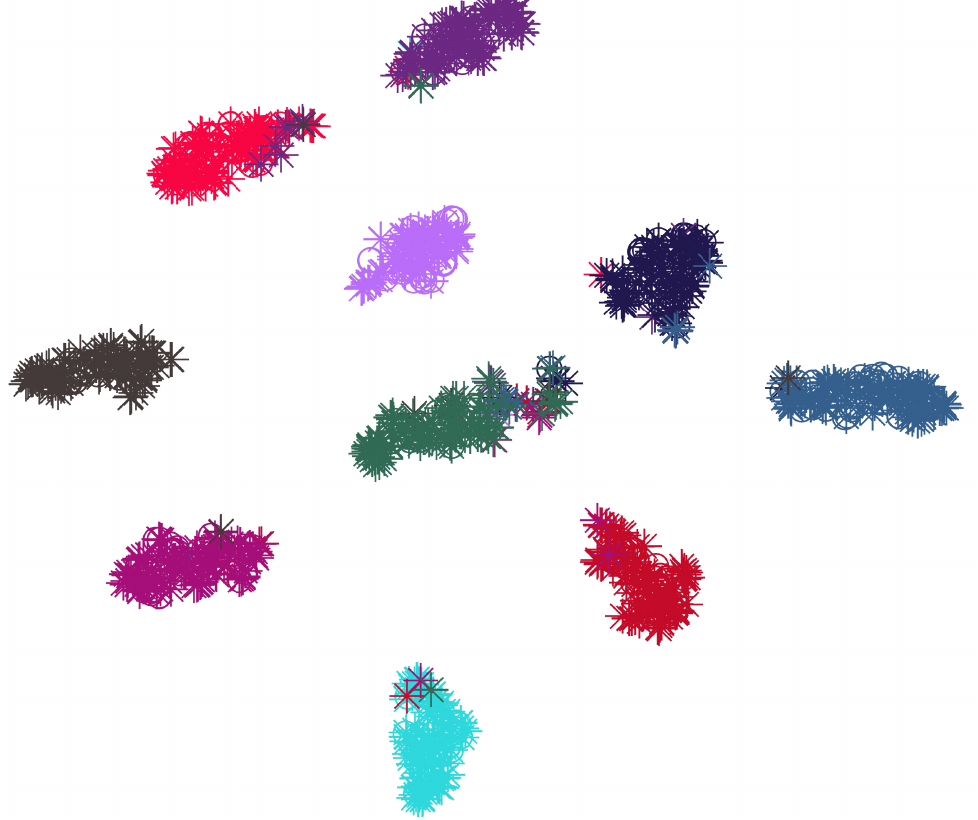}\label{AW3}}
	\caption{A t-SNE visualization of feature representations of different methods. Asterisk represents the features of Amazon(A), and circle represents the features of Webcam(W), where different colors represent different classes.}
	\label{visualization}
\end{figure*}
To verify the effectiveness of our landmark selection and manifold feature learning, we compare the experimental results of GGLS with and without landmark selection and manifold feature learning as shown in Fig.\ref{Nolandmark_NOmanifold}, where GGLS-noLS represents GGLS without landmark selection, GGLS-noMFL represents GGLS without manifold feature learning, and GGLS-noLSMFL represents GGLS without landmark selection and manifold feature learning. The results clearly indicate that our proposed landmark selection and manifold feature learning are all important and effective for transfer learning. GGLS-noLSMFL also can achieve comparable performance, but GGLS with landmark selection and manifold feature learning would reach better accuracies. {We further verify the effectiveness of kernel function by comparing GGLS-noLSMFL and GGLS-noLSMFLKF, where GGLS-noLSMFLKF represents GGLS-noLSMFL without kernel function. From Fig.\ref{Nolandmark_NOmanifold} we can observe that GGLS-noLSMFL achieves better performance than GGLS-noLSMFLKF. This indicates that it is more effective to explore the distribution of source domain and target domain in kernel space.}

The parameter $\mu$ balances the contributions of marginal distribution and conditional distribution. In order to evaluate the effectiveness of calculated $\mu$ by Eq.~(\ref{mu}), we compare our solution with $\mu=0.5$, which means the contribution of marginal distribution and conditional distribution are the same. The comparison is shown in Table\ref{compare_mu}, where the improvement shows the performance variation. For example, the improvement of mean accuracy on M$\rightarrow$ U is: $(90.6 - 89.2)/89.2 \times 100\% = 1.6\%$. From the table, we can see that the performance of our solution is better than $\mu=0.5$, which demonstrates the effectiveness in calculating $\mu$. Therefore, the importance of the two distributions is significantly different, which explains that the improvement is necessary.
\begin{table}[!htb]
	\centering
	\begin{tabular}{lrrr}
		\hline
		Data                 & $\mu$=0.5     & Estimated $\mu$   &  Improvement \\
		\hline
		A$\rightarrow$ C      &48.5 &48.6  &  0.2\%    \\
		A$\rightarrow$ D     &49.7   &51.0 &  2.6\%     \\
		C$\rightarrow$ A(Decaf$_{6}$)      & 94.3 &94.9  &  0.6\%     \\
		D$\rightarrow$ A(Decaf$_{6}$)      & 93.6 &94.1  &  0.5\%       \\
		M$\rightarrow$ U     &89.2   &90.6   &  1.6\%    \\
		V$\rightarrow$ I      & 77.5 &78.3    &  1.0\%   \\
		\hline
	\end{tabular}
	\caption{Comparison of $\mu$=0.5 and our estimated $\mu$.}
	\label{compare_mu}
\end{table}

\subsection{Visualization}
The feature representations of different methods can be investigated qualitatively. Therefore, we use t-SNE to visualize high-dimensional data by giving each datapoint a location in the projected 2D space. Fig.~\ref{visualization} shows the visualization of feature representations of task A$\rightarrow$W by different methods, where GGLS-noLS represents GGLS without landmark selection. It can be seen that the source and target are not aligned well with the Decaf feature, the samples from different classes are confused, and the inter-class distance is also small, which can easily lead to misclassification. They are aligned better and categories are discriminated better by GGLS-noLS, while GGLS is evidently better than GGLS-noLS. From Fig.~\ref{AW3} we can see that the intra-class samples are more compact and the
distributions between different classes are more dispersed. That is to say, GGLS can further reduce the distance of intra-class and increase the distance of inter-classes.
This shows the benefit of landmark selection.

\section{Conclusion}
\label{Conclusion}
In this paper, we propose a Grassmannian graph-attentional landmark selection (GGLS) framework for visual domain adaptation, which considers sample reweighting and feature matching simultaneously. GGLS learns an attention mechanism based sample reweighting, which treats the landmarks of each sample differently, and performs distribution adaptation and knowledge adaptation over the Grassmann manifold. Distribution adaptation balances the contributions of marginal and conditional distributions by our solution to estimate the balanced parameter. Knowledge adaptation preserves the local topology structure of samples and the discriminative power of samples from the source domain.
Extensive experiments have verified that GGLS significantly outperforms several state-of-the-art domain adaptation methods on three different real-world cross-domain visual recognition tasks.

\section*{Acknowledgments}
The work is supported by National Natural Science Foundation of China (Grant Nos. 61772049, 61632006, 61876012), and Scientific Research Common Program of Beijing Municipal Commission of Education (Grant Nos. KM201710005022, KM201510005024).

%\section*{References}

\bibliographystyle{unsrt}
\bibliography{mybibfile}

\end{document}